\documentclass[preprint,authoryear,11pt]{elsarticle}

\usepackage{geometry}
\usepackage{enumitem} 										
\setlist[itemize]{leftmargin = *}							
\setitemize{itemsep = 0 pt}									
			
\usepackage[utf8]{inputenc}
\usepackage[font=small,labelfont=bf]{caption}
\usepackage[T1]{fontenc}
\usepackage[english]{babel}	  
\usepackage{lmodern}
\usepackage{blindtext}

\usepackage{amsmath} 							
\usepackage{amsfonts}
\usepackage{amssymb}
\usepackage{mathtools}
\usepackage{delimset}

\usepackage{tabularx}
\usepackage{booktabs}

\usepackage{graphicx} 
\usepackage{caption}
\usepackage{algorithm}
\usepackage{algcompatible}	 
\algnewcommand\INPUT{\item[\textbf{Input:}]}		
\algnewcommand\OUTPUT{\item[\textbf{Output:}]}		

\usepackage{soul}
\usepackage{lineno}
\usepackage{lipsum}

\usepackage[explicit]{titlesec}

\usepackage{csquotes}
\usepackage{textcomp}
\usepackage{gensymb}
\usepackage{wrapfig}
\usepackage{mwe}
\usepackage{ragged2e}
\usepackage{stmaryrd}

\titleformat{\subsection}{\bfseries}{\thesubsection}{1em}{#1}	
\titlespacing*{\subsection}{0pt}{1ex plus .5ex minus .5ex}{.5em}

\titleformat{\paragraph}[runin]{\scshape}{\theparagraph}{.5em}{#1}	
\titlespacing*{\paragraph}{0pt}{1ex plus .5ex minus 0ex}{.5em}

\usepackage[usenames, dvipsnames]{xcolor}

\usepackage{url}
\usepackage[colorlinks]{hyperref}

\addto\extrasenglish{

}
\newcommand*{\secref}[1]{\hyperref[{#1}]{\autoref*{#1}. \nameref*{#1}}} 
\newcommand*{\defref}[1]{\hyperref[{#1}]{Definition \autoref*{#1}}}

\usepackage{amsthm}
\newtheoremstyle
{style}   									
{5pt}                     					
{5pt}                     					
{\normalfont}          						
{}                         					
{\bfseries}									
{.}  										
{ } 										
{}											
\theoremstyle{style}
\newtheorem{theorem}{Theorem}

\usepackage{amsthm}
\newtheoremstyle
{style}   									
{5pt}                     					
{5pt}                     					
{\normalfont}          						
{}                         					
{\bfseries} 								
{.}  										
{ } 										
{}											
\theoremstyle{style}
\newtheorem{definition}{Definition}

\newcommand{\vex}{\mbox{vec}}

\begin{document}

\hypersetup{colorlinks, linkcolor = Emerald, citecolor = Emerald}

\begin{frontmatter}
\title{An induction proof of the backpropagation algorithm in matrix notation}

\author{Dirk Ostwald, Franziska Usée}
\address{
Institute of Psychology and Center for Behavioral Brain Sciences, \\
Otto-von-Guericke Universität Magdeburg,
Germany}
\begin{abstract}
Backpropagation (BP) is a core component of the contemporary deep learning incarnation of neural networks. Briefly, BP is an algorithm that exploits the computational architecture of neural networks to efficiently evaluate the gradient of a cost function during neural network parameter optimization. The validity of BP rests on the application of a multivariate chain rule to the computational architecture of neural networks and their associated objective functions. Introductions to deep learning theory commonly present the computational architecture of neural networks in matrix form, but eschew a parallel formulation and justification of BP in the framework of matrix differential calculus. This entails several drawbacks for the theory and didactics of deep learning. In this work, we overcome these limitations by providing a full induction proof of the BP algorithm in matrix notation. Specifically, we situate the BP algorithm in the framework of matrix differential calculus, encompass affine-linear potential functions, prove the validity of the BP algorithm in inductive form, and exemplify the implementation of the matrix form BP algorithm in computer code.
\end{abstract}
\end{frontmatter}

\section{Introduction}
\thispagestyle{empty}
Backpropagation (BP) is a core component of the contemporary deep learning incarnation of neural networks \citep{lecun_deep_2015, schmidhuber_deep_2015}. In brief, BP is an algorithm that exploits the computational architecture of neural networks to efficiently evaluate the gradient of a cost function during gradient-based neural network parameter optimization. As reviewed by \citet{schmidhuber_deep_2015}, BP was developed and refined by multiple research groups during the 1970's and 1980's and popularized by \cite{rumelhart_learning_1986a}. Artificial intelligence and machine learning textbooks invariably feature tutorials on BP, so that BP has become firmly intertwined with deep learning theory and didactics \citep[e.g.,][]{haykin_neural_1998, duda_pattern_2001, bishop_pattern_2006, alpaydin_introduction_2014, nielsen_neural_2015, goodfellow_deep_2017, deisenroth_mathematics_2020}.

The validity of the BP algorithm rests on the application of a multivariate chain rule to the computational architecture of neural networks and their associated objective functions. As recently pointed out by \citet{mishachev_backpropagation_2017}, introductions to deep learning theory (such as those cited above) commonly present the computational architecture of neural networks in matrix form, but eschew a parallel formulation and justification of BP in the framework of matrix differential calculus. This entails several drawbacks for the theory and didactics of deep learning. First, intermingling matrix-based formulations of the neural network forward architecture with coordinate-based formulations of the ensuing BP algorithm results in a mathematically unsatisfying representation of core deep learning components. Second, shuffling together different levels of mathematical granularity (i.e., matrix-based forward pass formulations and coordinate-based BP update equations) renders the fundamental theory of deep learning unnecessarily opaque, especially for novices in the field. Third, didactic implementations of neural network training in array-based environments are hampered by absent formulations and insufficient justifications of BP procedures.

In this article, we aim to overcome these limitations in the formulation of core deep learning components by providing a full induction proof of the BP algorithm in matrix notation. In our work, we go beyond the account of \citet{mishachev_backpropagation_2017} by, first, formally situating the BP algorithm in the framework of matrix differential calculus  \citep[][see\autoref{sec:appendix_1} for an overview of required concepts]{magnus_matrix_1989}, second, explicitly encompassing affine-linear potential functions rather than focussing on  homogeneous (bias-free) neural networks, third, proving the validity of the BP algorithm in explicit inductive form, and fourth, providing an implementation of the matrix form BP algorithm with the software that accompanies this article. Taken together, we thus provide a novel formal grounding of a core component of contemporary deep learning that may not only serve as a didactic resource in the training of aspiring data scientists, but may also inspire the exploration of novel matrix analysis-based approaches in neural network training.

The outline of this article is as follows. In Section \secref{sec:neural_networks}, we first review the essential building blocks of neural networks. This section primarily serves to introduce the notational conventions that apply in this article. Section \secref{sec:neural_network_training} then sets the scene for introducing the BP algorithm by introducing gradient descent as a method for learning neural network parameters based on training data. Sections \secref{sec:BP_algorithm} and\secref{sec:proof} form the core of the article and provide the inductive proof of the validity of the BP algorithm for computing a neural network's cost function gradient. The mathematical background for these Sections is provided in \secref{sec:appendix_1}, \secref{sec:appendix_2}, and \secref{sec:appendix_3}. Finally, we discuss an exemplary application of the BP algorithm in matrix form in Section \secref{sec:exemplary_application}. The Matlab implementation of this application, as well as all code generating the figures of this article is available at \url{https://osf.io/7awpj/}. 

A few remarks on the notational conventions used in this article are in order. For conformable matrices $A$ and $B$, we denote the standard matrix product by $A \cdot B$, the Kronecker matrix product by $A \otimes B$, and the Hadamard matrix product by $A \circ B$. We denote the $n \times n$ diagonal matrix formed by the components of a vector $v \in \mathbb{R}^n$ by $\mbox{diag}(v)$. For a matrix $A \in \mathbb{R}^{m \times n}$, we use $A_\bullet$ to denote the $m \times n-1$ matrix that results from removing the last column of $A$.

\section{Neural networks}\label{sec:neural_networks}
\subsection{Basic definitions}
Neural networks are generally conceived as parameterized multivariate vector-valued functions that are characterized by a serial concatenation of affine-linear and nonlinear function dyads. In lack of a generally accepted term, the affine-linear functions are here referred to as \textit{potential functions}. Potential functions are parameter-dependent and the parameters of all potential functions constitute a neural networks' parameter set. The nonlinear functions are generally referred to as \textit{activation functions} and do not have adjustable parameters. We make these concepts precise in \defref{def:potential_functions}, \defref{def:activation_functions}, and \defref{def:neural_network} and subsequently discuss their relation to the common neural network nomenclature. 

\begin{definition}[Potential functions]\label{def:potential_functions}
\justifying
Let $W \in \mathbb{R}^{m \times (n+1)}$ and $a \in \mathbb{R}^n$, which we shall refer to as a \textit{weight matrix} and an \textit{activation vector}, respectively. We call a function 
of the form
\begin{equation}\label{eq:Phi}
\Phi : \mathbb{R}^{m \times (n+1)} \times \mathbb{R}^{n} \to \mathbb{R}^m, 
(W,a) \mapsto \Phi(W,a) 
:= W\cdot
\begin{pmatrix}
a \\ 1
\end{pmatrix}  
\end{equation} 
a \textit{bivariate potential function}. For fixed $a$, we call a function of the form
\begin{equation}\label{eq:Phi_a}
\Phi_a : \mathbb{R}^{m \times (n+1)} \to \mathbb{R}^m, \, 
W \mapsto \Phi_a(W) := \Phi(W,a)
\end{equation}
a \textit{weight matrix-variate potential function}, whereas for fixed $W$, we call a function of the form 
\begin{equation}\label{eq:Phi_W}
\Phi_W : \mathbb{R}^{n} \to \mathbb{R}^m, 
a \mapsto \Phi_W(a) := \Phi(W,a)
\end{equation}
a \textit{potential function}. Finally, we call $z := \Phi_W(a)$ a \textit{potential vector}.

$\hfill \bullet$
\end{definition}

\begin{table}[t]
\begin{center}
\renewcommand{\arraystretch}{2}
\begin{small}
\begin{tabularx}{\textwidth}{XXX}
\textbf{Denomination}
& \textbf{Definition}
& \textbf{Derivative}
\\ 
\hline
Standard logistic 
& $\sigma(z_i) := \frac{1}{1+\exp(-z_i)}$ 
& $\sigma'(z_i) = \frac{\exp(z_i)}{(1 + \exp(z_i))^2}$
\\ 
Hyperbolic tangent  
& $\sigma(z_i) := \mbox{tanh}(z_i)$ 
& $\sigma'(z_i) = 1 - \mbox{tanh}^2(z_i)$ 
\\ 
ReLU 
& $\sigma(z_i) := \max(0, z_i)$	 
& $\sigma'(z_i) = 
\begin{cases}
0, 					& z_i < 0 \\
\emptyset,			& z_i = 0 \\
1, 					& z_i > 0 \\
\end{cases}$
\\ 
Leaky ReLU 
& 
$\sigma(z_i) := 
\begin{cases}
0.1z_i,	& z_i \le 0 	\\
z_i, 		& z_i > 0 	\\
\end{cases}$
& $\sigma'(z_i) = 
\begin{cases}
0.01,		& z_i \le 0 	\\
1, 			& z_i > 0 	\\
\end{cases}$
\\ 
\end{tabularx} 
\end{small}
\end{center}
\caption{Commonly used activation functions and their derivatives.}\label{tab:table_1}
\end{table}
\vspace{5mm}

\begin{figure}[t]
\begin{center}
\includegraphics[width = .95\textwidth]{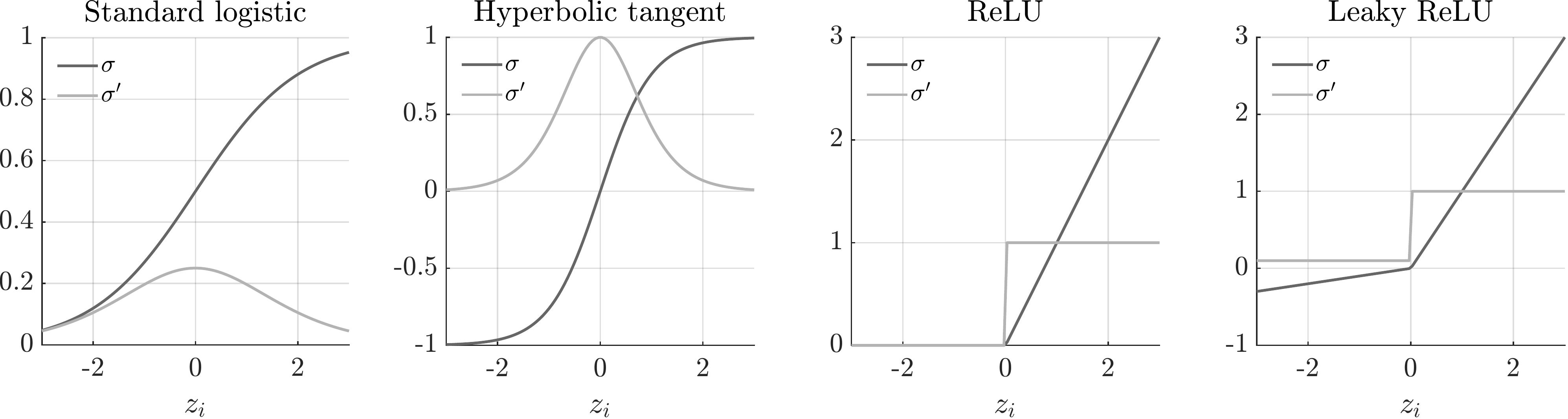}
\end{center}
\caption{Graphs of the commonly used activation functions and their derivatives as listed in \autoref{tab:table_1}.}\label{fig:figure_1}
\end{figure}

\begin{definition}[Activation functions]\label{def:activation_functions}
A multivariate vector-valued function
\begin{equation}\label{eq:Sigma}
\Sigma : \mathbb{R}^n \to \mathbb{R}^n, 
z \mapsto \Sigma(z) := (\sigma(z_1), ..., \sigma(z_n))^T,
\end{equation} 
where
\begin{equation}\label{eq:sigma}
\sigma: \mathbb{R} \to \mathbb{R},  
z_i \mapsto \sigma(z_i) =: a_i 
\mbox{ for all } i = 1,...,n,
\end{equation}
is called a \textit{component-wise activation function} and the univariate real-valued function $\sigma$ is called an \textit{activation function}. 

$\hfill \bullet$
\end{definition}

\noindent Commonly used activation functions and their derivatives are listed in \autoref{tab:table_1} and are visualized in \autoref{fig:figure_1}. Based on the concepts of potential functions and component-wise activation functions, a $k$-layered neural network can be defined as follows (cf. \citet[][Section 2]{mishachev_backpropagation_2017}):
\begin{definition}[$k$-layered neural network]\label{def:neural_network}
A multivariate vector-valued function 
\begin{equation}
f : \mathbb{R}^{n_0} \to \mathbb{R}^{n_k}, x \mapsto f(x) =: y 
\end{equation}
is called a \textit{$k$-layered neural network}, if $f$ is of the form
\begin{equation}
\footnotesize
f : \mathbb{R}^{n_0}
\xrightarrow[]{\Phi^1_{W^1}}
\mathbb{R}^{n_1}
\xrightarrow[]{\Sigma^1}
\mathbb{R}^{n_1}
\xrightarrow[]{\Phi^2_{W^2}}
\mathbb{R}^{n_2}
\xrightarrow[]{\Sigma^2}
\mathbb{R}^{n_2}
\xrightarrow[]{\Phi^3_{W^3}}
\cdots 
\xrightarrow[]{\Phi^{k-1}_{W^{k-1}}}
\mathbb{R}^{n_{k-1}}
\xrightarrow[]{\Sigma^{k-1}}
\mathbb{R}^{n_{k-1}}
\xrightarrow[]{\Phi^k_{W^k}}
\mathbb{R}^{n_{k}}
\xrightarrow[]{\Sigma^{k}}
\mathbb{R}^{n_{k}},
\end{equation}
where for  $l = 1,...,k$
\begin{equation}
\Phi^l_{W^l} : \mathbb{R}^{n_{l-1}} \to \mathbb{R}^{n_l}, 
a^{l-1} \mapsto \Phi^l_{W^l}(a^{l-1}) := 
W^l 
\cdot
\begin{pmatrix}
a^{l-1} \\ 1
\end{pmatrix}
=: z^l
\end{equation}
are potential functions and 
\begin{equation}
\Sigma^l : \mathbb{R}^{n_l} \to \mathbb{R}^{n_l}, 
z^l \to \Sigma^l(z^l) =: a^l
\end{equation}
are component-wise activation functions. For $x \in \mathbb{R}^{n_0}$, a $k$-layered neural network takes on the value
\begin{equation}
f(x) := 
\Sigma^{k}
(\Phi^k_{W^k}(\Sigma^{k-1}(\Phi^{k-1}_{W^{k-1}}(\Sigma^{k-2}(\cdots(\Sigma^{1}(\Phi^1_{W^1}(x))\cdots)))))) 
\in \mathbb{R}^{n_k}.
\end{equation}
$\hfill \bullet$
\end{definition}

\noindent The common nomenclature associated with the constituents of a homogeneous $k$-layered neural network is as follows. 
\begin{itemize}[leftmargin=*]
\justifying
\item the vectors $a^l = (a_1^l, ..., a_{n_l}^l)^T \in \mathbb{R}^{n_l}$ for $l = 0,1,..., k$ are called \textit{activation vectors of layer $l$}, 
\item the components $a_i^l \in \mathbb{R},i = 1,...,n_l, l = 0,1,...,k$ are called \textit{neuron activations of layer $l$},
\item the layer with index $l = 0$ and dimension $n_0$ is called the \textit{network input layer},
\item the activation vector with index $l = 0$ is called \textit{input} and is denoted by $x := a^0$,
\item the layer with index $l = k$ and dimension $n_k$ is called the \textit{network output layer},
\item the activation vector with index $l = k$ is called \textit{output} and is denoted by $y := a^k$, and 
\item the layers with indices $l = 1,...,k-1$ are called \textit{hidden layers}.
\end{itemize}
At the level of individual neural network neurons, the following nomenclature ensues: let $w_{ij}^l \in \mathbb{R}$ denote the $ij$th entry in the $l$th weight matrix, i.e., 
\begin{equation}
W^l = (w_{ij}^l)_{1 \le i \le n_l, 1 \le j \le n_{l-1}+1} \in \mathbb{R}^{n_l \times (n_{l-1}+1)}
\mbox{ for }
l = 1,...,k.
\end{equation}
Then for $i = 1,...,n_l$ and $j = 1,...,n_{l-1}$, $w_{ij}^l$ is the \textit{synaptic weight} connecting neuron $j$ in layer $l-1$ and neuron $i$ in layer $l$, while for $i = 1,...,n_l$, $w_{i,n_{l-1}+1}$ is the \textit{bias} of neuron $i$ in layer $l$. For $l = 1,..., k$, the last column of $W^l$ thus encodes the biases of the neurons in layer $l$. The \textit{potential} of neuron $i$ in layer $l$ for $i = 1,...,n_l$ and $l = 1,...,k$  is given by
\begin{equation}
z_i^l = \sum_{j = 1}^{n_{l-1}} w^l_{ij}  a_j^{l-1} + w_{i, n_{l-1} + 1} \in \mathbb{R}^{n_l}.
\end{equation}
Finally, based on the functional form of the component-wise activation function, the activation of neuron $i$ in layer $l$ for $i = 1,...,n_l$ and $l = 1,...,k$ is given by
\begin{equation}
a_i^l = \sigma\left(\sum_{j = 1}^{n_{l-1}} w^l_{ij}  a_j^{l-1} + w_{i, n_{l-1} + 1} \right) \in \mathbb{R}^{n_l}, 
\end{equation}
and may be conceived as the \textit{mean firing rate} of the $i$th neuron in layer $l$. 

\begin{figure}[t!]
\begin{center}
\includegraphics[width = .9\textwidth]{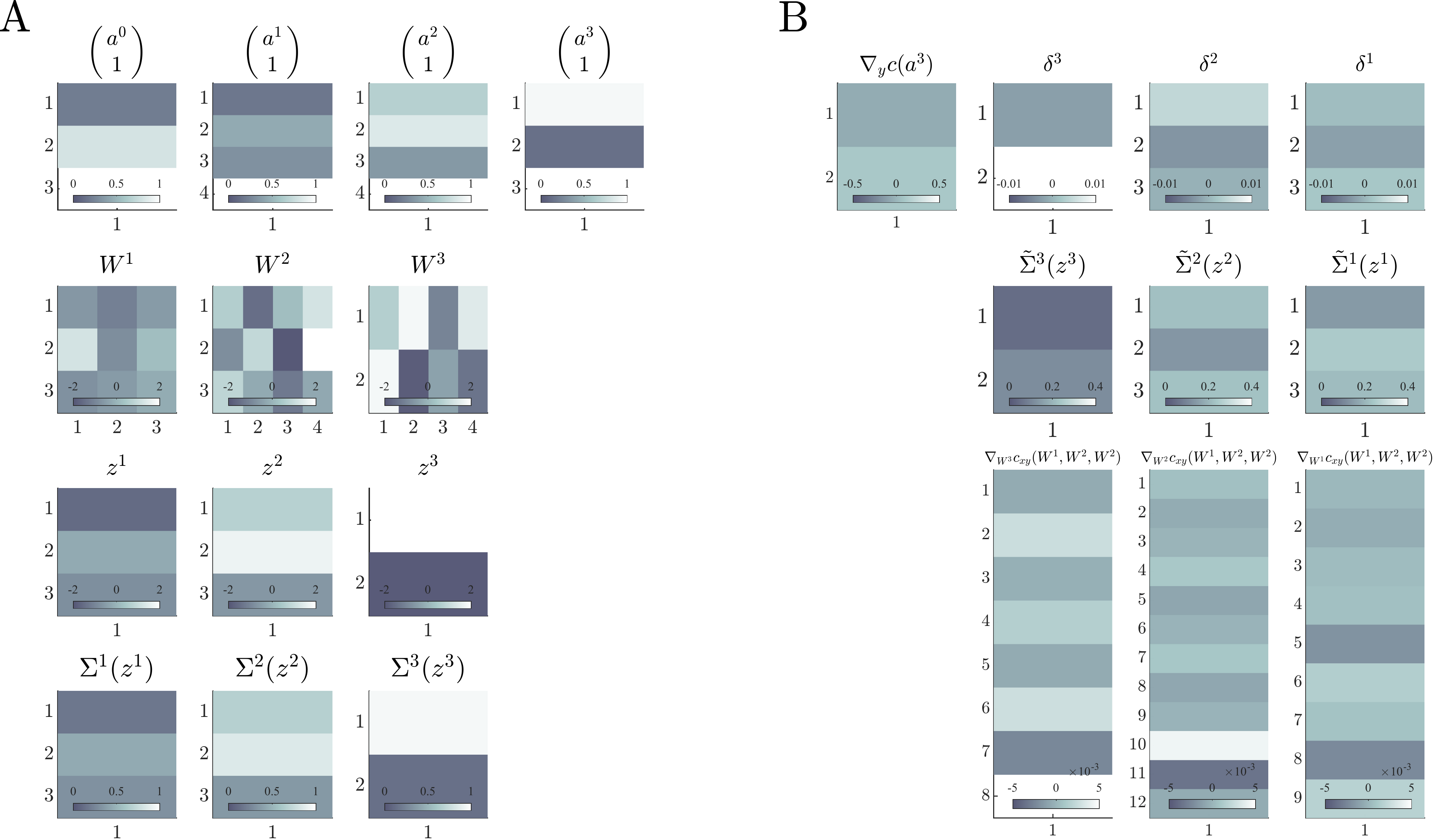}
\caption{\textbf{(A)} Exemplary forward pass evaluation for a 3-layered neural network as specified in Example 1. In descending order, the panel rows depict (1) the augmented activation vectors of layers $l = 0$ to $l = 3$ for the input $x = (0.2,0.8)^T$, (2) the weight matrices $W^1, W^2, W^3$, the elements of which were sampled independently and identically distributed from a univariate Gaussian distribution with expectation parameter 0 and variance parameter 1, (3) the weighted inputs $z^1, z^2,z^3$, and (4) the results of the application of the component-wise activation functions with a logistic activation function to each weighted input, $\Sigma^{(1)}(z^1),\Sigma^{(2)}(z^2),\Sigma^{(3)}(z^3)$. Note that the components of $\Sigma^{(l)}(z^l)$ for $l = 1,2,3$ are lower and upper bounded by 0 and 1, respectively, and that $\Sigma^{(l)}(z^l)  = a^l$ for $l = 1,2,3$. \textbf{(B)} Exemplary BP algorithm evaluation for the 3-layered neural network as specified in Example 1 for the training exemplar $((0.2,0.8)^T, (1,0)^T)$ and with weight matrices set as visualized in (A). In descending order, the panel rows depict (1) the gradient of the output-specific cost function, corresponding to $\delta^4$, as well as the BP error vectors $\delta^3, \delta^2, \delta^1$, (2) the vector format-form of the component-wise activation function derivative $\tilde{\Sigma}^3(z^3), \tilde{\Sigma}^2(z^2)$, $\tilde{\Sigma}^3(z^3)$, and (3) the partial gradients of the training exemplar-specific cost function with respect to $W^3$, $W^2$, and $W^1$.}
\label{fig:figure_2}
\end{center}
\end{figure}

\paragraph{Example 1.}\label{sec:example_1} To illustrate the definitions above, we specify the key components of a 3-layered neural network ($k = 3$) with a two-dimensional input layer ($n_0 = 2$), two three-dimensional hidden layers ($n_1 = 3$, $n_2 = 3$), and a two-dimensional output layer ($n_3 = 2$) below. Note that $x = a^0$ and $a^3 = y$. Exemplary constituents of this neural network for a given input $x \in \mathbb{R}^2$ are visualized in \autoref{fig:figure_2}A. 

\begin{center}
\begin{footnotesize}
\renewcommand{\arraystretch}{1.1}
\begin{tabular}{llllll}
$\begin{pmatrix}
a^0 \\
1
\end{pmatrix}$
&
$ = \begin{pmatrix}
a_1^0 \\
a_2^0 \\
1
\end{pmatrix}$ 
& 
$\begin{pmatrix}
a^1 \\
1
\end{pmatrix}$
&
$ = \begin{pmatrix}
a_1^1 \\
a_2^1 \\
a_3^1 \\
1 	    	
\end{pmatrix}$ 
&
$\begin{pmatrix}
a^2 \\ 1
\end{pmatrix}$
&
$ =
\begin{pmatrix}
a_1^2 \\
a_2^2 \\
a_3^2 \\
1
\end{pmatrix}$
\quad\quad\quad
$a^3
=
\begin{pmatrix}
a_1^3 \\
a_2^3 \\
\end{pmatrix}$ 
\\
\\
$W^1$
& 
$ = \begin{pmatrix}
w_{11}^1 & w_{12}^1 & w_{13}^1 \\
w_{21}^1 & w_{22}^1 & w_{23}^1 \\
w_{31}^1 & w_{32}^1 & w_{33}^1 \\
\end{pmatrix}$  
& 
$W^2$ 
&
$ = \begin{pmatrix}
w_{11}^2 & w_{12}^2 & w_{13}^2  & w_{14}^2 	\\
w_{21}^2 & w_{22}^2 & w_{23}^2  & w_{24}^2	\\
w_{31}^2 & w_{32}^2 & w_{33}^2  & w_{34}^2 	\\
\end{pmatrix}$ 
&
$W^3$ 
& 
$= \begin{pmatrix}
w_{11}^3 & w_{12}^3 & w_{13}^3 & w_{14}^3 \\
w_{21}^3 & w_{22}^3 & w_{23}^3 & w_{24}^3 \\
\end{pmatrix}$  
\\
\\
$z^1$ 
&
$ = \begin{pmatrix}
z_{1}^1 \\
z_{2}^1 \\
z_{3}^1
\end{pmatrix}$ 
&
$z^2$ 
&
$ = \begin{pmatrix}
z_{1}^2 \\
z_{2}^2 \\
z_{3}^2 
\end{pmatrix}$
& 
$z^3$ 
& = 
$\begin{pmatrix}
z_{1}^3 \\
z_{2}^3 \\
\end{pmatrix}$  
\\
\\
$\Sigma^1(z^1)$ 
&
$ = \begin{pmatrix}
\sigma(z_1^1) 	\\
\sigma(z_2^1) 	\\
\sigma(z_3^1)	\\
\end{pmatrix}$
&
$\Sigma^2(z^2)$
&
$ =\begin{pmatrix}
\sigma(z_1^2) \\
\sigma(z_2^2) \\
\sigma(z_3^2) \\
\end{pmatrix}$ 
&
$\Sigma^3(z^3)$ 
&
$ =
\begin{pmatrix}
\sigma(z_1^3) \\
\sigma(z_2^3) \\
\end{pmatrix}$
\end{tabular} 
\end{footnotesize}
\end{center}

$\hfill \bullet$

\defref{def:neural_network} conceives a neural network as a function of an input $x$ for a given set of fixed weight matrices $W^l,l = 1,...,k$. Neural network training, however, requires monitoring the output of a neural network for fixed input $x$ as a function of variable weight matrices $W^l, l = 1,...,k$. To formalize this fundamental change of perspective, we use the following definitions.
\begin{definition}[Weight matrix-variate neural network functions]
\label{def:matrix_variate_neural_network_function}
Let $f$ denote a $k$-layered neural network and let $x$ denote an input of $f$. Then the \textit{weight matrix-variate neural network function $f_x$ of $f$} is defined as the function
\begin{multline}\label{eq:f_x}
f_x : \mathbb{R}^{n_{1} \times (n_0 + 1)} \times \cdots \times \mathbb{R}^{n_{k} \times (n_{k-1} + 1)} 
\to \mathbb{R}^{n_k}, 
(W^1, ..., W^k) \mapsto f_x(W^1,...,W^k)
\\
:= \Sigma^k(\Phi^k(W^k,\Sigma^{k-1}(\Phi^{k-1}(W^{k-1}, \cdots (W^2,\Sigma^1(\Phi^1(W^1,x)))\cdots)))),
\end{multline}
where for $l = 1,...,k$, $\Phi^l$ denotes the bivariate potential function corresponding to the potential function $\Phi^l_{W^l}$ in the neural network's definition. Furthermore, for $l = 1,..., k$, we define the \textit{$l$th layer weight matrix-variate neural network function $f_x^l$ of $f$} for fixed $W^\ell \in \mathbb{R}^{n_\ell \times (n_{\ell-1} + 1)}$ with $\ell = 1,..., k$ and $\ell \neq l$ as
\begin{equation}\label{eq:f_x_l}
f_x^l : \mathbb{R}^{n_{1} \times (n_{l-1} + 1)} \to \mathbb{R}^{n_k}, 
W^l \mapsto f_x^l(W^l) := f_x^l(W^1, ..., W^k).
\end{equation}
$\hfill \bullet$
\end{definition}  
\noindent Note that the functional definition of $f_x$ in eq. \eqref{eq:f_x} is conceived as a function of the weight matrices $W^1, ..., W^l$ only.

\section{Neural network training}\label{sec:neural_network_training}
Neural network training is the process of adjusting a neural network's weight parameters to minimize some criterion of deviation between the neural network's outputs across a set of training input examples and their associated training output examples. The criterion of deviation is expressed by means of a cost function. In the following, we first formalize the notions of training sets, neural network training, and cost functions and then formulate a neural network gradient descent algorithm. 

\begin{definition}[Neural network training set] 
A \textit{neural network training set} is a set of vector pairs
\begin{equation}
\mathcal{D}
:= \lbrace(x^{(i)}, y^{(i)})\rbrace_{i=1}^n,
\end{equation}
where $x^{(i)} \in \mathbb{R}^{n_0}$ is referred to as \textit{feature vector} and $y^{(i)} \in \mathbb{R}^{n_k}$ is referred to as \textit{target vector}.

$\hfill \bullet$
\end{definition}
\noindent Typical target vector formats in neural network applications include $y^{(i)} \in \{0,1\}$ for binary classification problems, $y^{(i)} \in \{0,1\}^{n_k}$ with $\sum_{i=1}^{n_k} y_i = 1, n_k > 1$ for $n_k$-fold classification using a ,,one-hot-encoding'' scheme, and $y^{(i)} \in \mathbb{R}^{n_k}, n_k > 1$ for neural network regression. Given a neural network training set, neural network training is defined as follows.
\begin{definition}[Neural network training]
Let $f$ denote a $k$-layered neural network and let $\mathcal{D}$ denote a neural network training set. Then \textit{neural network training} is the process of adapting the neural network's weight matrices $W^1, ..., W^k$ with the aim of minimizing a deviation criterion between the neural network's output layer activation $f(x^{(i)})$ and the associated value of the target vector $y^{(i)}$ across all training exemplars $(x^{(i)},y^{(i)}), i = 1,...,n$ of a training set $\mathcal{D}$. 

$\hfill \bullet$
\end{definition}
\noindent As mentioned above, the deviation criterion is typically formalized in terms of cost functions. We first define the notions of \textit{output-specific} and \textit{training exemplar-specific} cost functions.

\begin{figure}[t]
\begin{center}
\includegraphics[width = .75\textwidth]{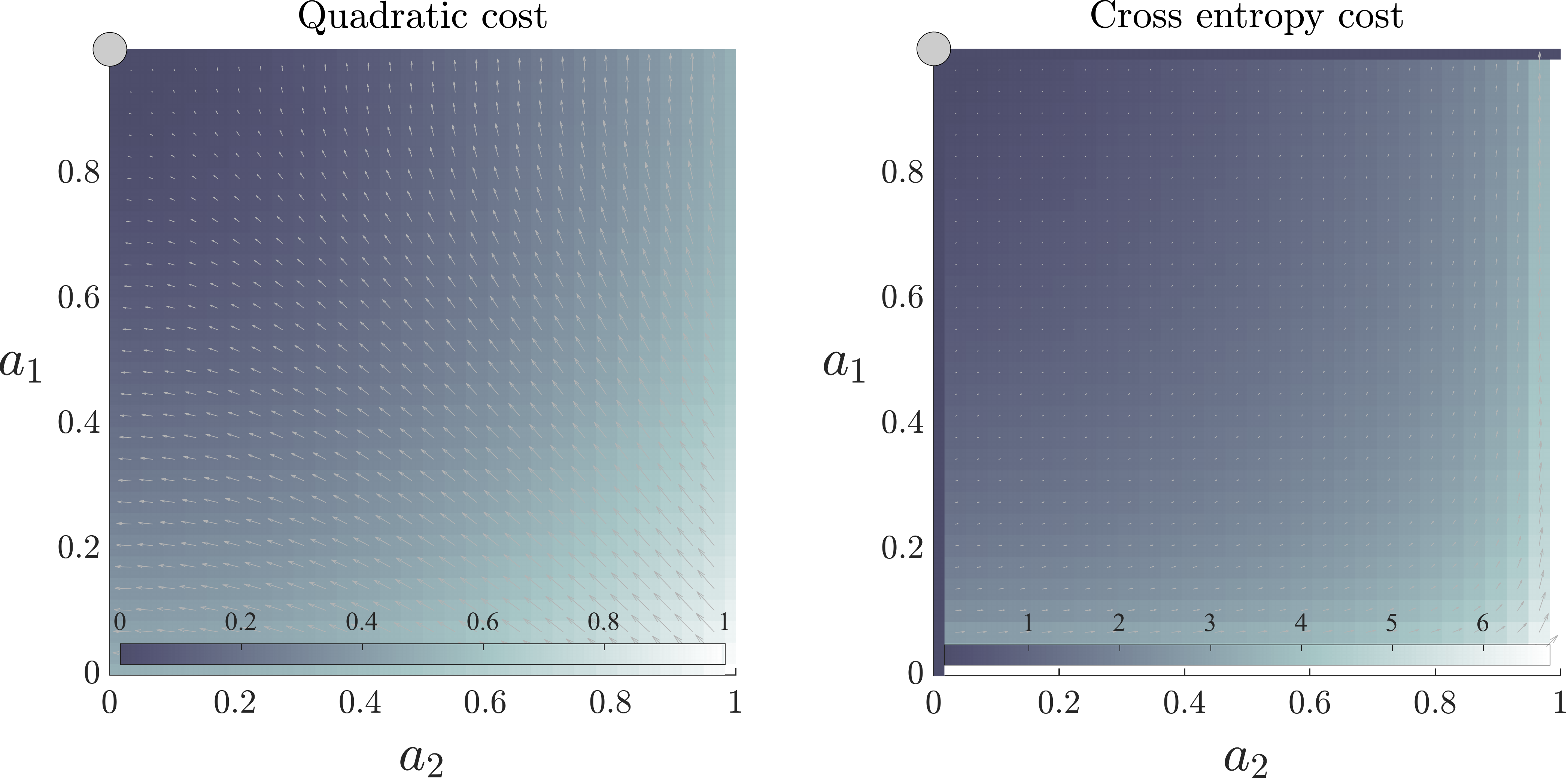}
\end{center}
\caption{Visualizations of the commonly used output-specific cost functions and their gradients listed in \autoref{tab:table_2}. The subpanels depict the respective output-speicific cost function values for $y = (1,0)^T$ as indicated by the light gray dot as a function of the neural network's output activation $a = (a_1,a_2)^T$ for a neural network with logistic activation function. Note that for such a network $(a_1,a_2)^T \in ]0,1[^2$ and thus the cross entropy cost function is well-defined. Both cost functions exhibit their minimum at $a = y$. The arrows indicate the respective function's gradient values, scaled appropriately for visualization purposes.}\label{fig:figure_3}
\end{figure}

\begin{definition}[Output-specific cost functions, training exemplar-specific cost functions]\label{def:cost_functions}
Let $f$ denote a $k$-layered neural network, let $f_x$ denote its associated weight matrix-variate neural network function, let $x$ and $y$ denote neural network inputs and outputs, respectively, and let $\mathcal{D}$ denote a neural network training set. Then a multivariate real-valued function of the form  
\begin{equation}\label{eq:output_cost_function}
c_y : \mathbb{R}^{n_k} \to \mathbb{R}, a^k \mapsto c_y(a^k)
\end{equation}
is called an \textit{output-specific cost function}. Furthermore, a multi-matrix-variate real-valued function of the form 
\begin{align}\label{eq:training_exemplar_cost_function}
\footnotesize
\begin{split}
c_{xy} : 
\mathbb{R}^{n_1 \times (n_0 + 1)} \times \cdots \times
\mathbb{R}^{n_k \times (n_{k-1} + 1)} 
\to \mathbb{R},\,
(W^1, ..., W^k) \mapsto c_{xy}(W^1, ..., W^k) := c_y(f_x(W^1,...,W^k))
\end{split}
\end{align}
is called \textit{training exemplar-specific cost function}. 

$\hfill \bullet$
\end{definition}
\noindent Commonly employed output-specific loss functions, the \textit{quadratic loss function} and the \textit{cross-entropy loss function}, as well as their gradients with respect to $a$, are listed in \autoref{tab:table_2} and visualized for a two-dimensional output in \autoref{fig:figure_3}. 

\begin{table}[t]
\begin{center}
\renewcommand{\arraystretch}{2}
\begin{small}
\begin{tabularx}{\textwidth}{lll}
\textbf{Denomination}
& \textbf{Definition}
& \textbf{Gradient}
\\ 
\hline
Quadratic cost
& $c_y(a^k) := \frac{1}{2}\sum_{j=1}^{n_k}(a_j^k - y_j)^2$ 
& $\nabla c_y(a^k) := \left(a_j^k - y_j\right)_{j = 1,...,n_k}$ 
\\ 
Cross entropy cost  
& $c_y(a^k) := -\sum_{i=1}^{n_k}y_j \ln a_j^k + (1 - y_j)\ln (1-a_j^k)$ 
& $\nabla c_y(a^k) := \left(-\frac{y_j}{a_j^k} + \frac{1-y_j}{1-a_j^k}\right)_{j = 1,...,n_k}$  
\end{tabularx} 
\end{small}
\end{center}
\caption{Commonly employed output-specific cost functions and their gradients.}\label{tab:table_2}
\end{table}
\vspace{1cm}

Neural network training typically proceeds by adjusting the neural network's weight vector such as to minimize an additive cost function. We define these concepts as follows.
\begin{definition}[Weight vector, additive cost function]\label{def:additive_cost_function}
Let $f$ denote a $k$-layered neural network with $n_l \times n_{l-1} + 1$-dimensional weight matrices $W^l, l = 1,.., k$ and let $p:=\sum_{l=1}^{n_k} n_{l} (n_{l-1} + 1)$. Then 
\begin{equation}\label{eq:weight_vector}
\mathcal{W} := 
\begin{pmatrix}
\vex\left(W^l\right)
\end{pmatrix}_{1 \le l \le k}
\in \mathbb{R}^p
\end{equation}
is called the neural network's \textit{weight vector}. Furthermore, a multivariate real-valued function of the form
\begin{equation}\label{eq:additive_cost_function}
c_{\mathcal{D}}: \mathbb{R}^{p} \to \mathbb{R},
\mathcal{W} \mapsto c_\mathcal{D}(\mathcal{W})
:= \frac{1}{n}\sum_{i=1}^n c_{x^{(i)}y^{(i)}}(W^1, ..., W^k)
\end{equation}
is called an \textit{additive cost function}.

$\hfill \bullet$
\end{definition}
\noindent Note that in \defref{def:additive_cost_function} an additive cost function is defined as a multivariate real-valued function, while a training exemplar-specific cost function is defined in \defref{def:cost_functions} as a multi-matrix-variate real-valued function. We thus tacitly assume the appropriate rearrangement of the weight vector $\mathcal{W}$ into the weight matrices $W^1, ..., W^k$ in the evaluation of the $c_\mathcal{D}$. We are now in the position to formulate the standard gradient descent algorithm for minimization of a neural network's additive cost function.

\begin{definition}[Neural network gradient descent algorithm]\label{def:gradient_descent}
Let $f$ denote a $k$-layered neural network with weight vector $\mathcal{W}$, let $\mathcal{D}$ denote a neural network training set comprising $n$ training exemplars, and let $c_{\mathcal{D}}$ denote an additive neural network cost function with associated training exemplar-specific cost function $c_{xy}$. Then a \textit{neural network gradient descent algorithm} is an algorithm of the following form.

\vspace{1mm}

\noindent\underline{Initialization}
\vspace{1mm}

\noindent Select $\mathcal{W}^{(0)}$ and $\alpha > 0$ appropriately.
\vspace{1mm}

\noindent\underline{Iterations}
\vspace{1mm}

\noindent For $j = 1,2,...$ until convergence, set
\begin{equation}\label{eq:w_update}
\mathcal{W}^{(j)} := 
\mathcal{W}^{(j-1)} 
- \frac{\alpha}{n}
\sum_{i=1}^n 
\nabla c_{x^{(i)}y^{(i)}}(W^1, ..., W^k),
\end{equation}
where 
\begin{equation}\label{eq:c_gradient}
\nabla c_{x^{(i)}y^{(i)}}(W^1, ..., W^k)
= 
\begin{pmatrix}
\nabla_{W^l} c_{x^{(i)}y^{(i)}}(W^1, ..., W^k)	
\end{pmatrix}_{1\le l \le k}
\end{equation} 
denotes the gradient of the $i$th training exemplar-specific cost function for $i = 1,...,n$.

$\hfill \bullet$
\end{definition}
\noindent Note that by eq. \eqref{eq:w_update}, $\mathcal{W}^{(j)}$ is adapted in the negative average gradient direction over \textit{all} training exemplars, a procedure known as \textit{batch gradient descent}.  If the average is formed over a (randomly) selected subset of training exemplars only, the neural network gradient descent algorithm is referred to as \textit{stochastic gradient descent}. 

\section{The BP algorithm in matrix notation}\label{sec:BP_algorithm}
The BP algorithm is a computational procedure to evaluate the partial gradients of training exemplar-specific neural network cost function that constitute that function's gradient (cf. eq. \eqref{eq:c_gradient}). A matrix version of the BP algorithm for homogeneous neural networks was recently given by \citet[][Section 5]{mishachev_backpropagation_2017}. We state the matrix version of the BP algorithm for neural networks in theorem form below and show its validity in \secref{sec:proof}.

\begin{theorem}[Backpropagation algorithm]\label{thm:BP_algorithm}
Let $f$ denote a $k$-layered neural network, let $W_\bullet^{l} \in \mathbb{R}^{n_l \times n_{l-1}}$ for $l = 1,...,k$ denote matrices formed by removing the last column of the neural network's weight matrices $W^{l} \in \mathbb{R}^{n_l \times n_{l-1} + 1}$, let $c_{xy}$ denote a training exemplar-specific cost function, let $\nabla c_y(a^k)$ denote the gradient of the output-specific cost function, let $\tilde{\Sigma}^l(z^l) := (\sigma'(z^l_1), ..., \sigma'(z^l_{n_l}))^T$ denote the vector of activation function derivatives evaluated at $z^l$, and let $\Sigma^l(z^l)$ denote component-wise activation functions evaluated at $z^l$. Then the partial gradients of $c_{xy}$ with respect to the weight matrices $W^l$ for $l = k, k-1, ...,1$ can be computed according to the following algorithm: 
\vspace{1mm}

\noindent\underline{Initialization}
\vspace{1mm}

\noindent Set 
$W^{k+1} := 
(\begin{matrix}
1 & 0
\end{matrix})$ 
and 
$\delta^{k+1} := \nabla c_y(a^{k})$.
\vspace{1mm}

\noindent \underline{Iterations}
\vspace{1mm}

\noindent For $l = k, k-1, k-2, ..., 1$, set
\begin{equation}\label{eq:error}
\delta^l 
:= 	\brk{
		 (W^{l+1}_{\bullet})^T \cdot \delta^{l+1}
		} 
	\circ \tilde{\Sigma}^{l}(z^l) 
\end{equation}
\noindent and
\begin{equation}\label{eq:partial_gradient}
\nabla_{W^l} 
c_{xy}(W^1,..., W^k) 
:= \vex 
	\brk{
		\delta^l 
		\cdot 
		\brk{\begin{matrix} \Sigma^{l-1} (z^{l-1})^T & 1 \end{matrix}}
		},
\end{equation}
where the recursion is terminated by 
$\Sigma^0(z^0) := x^T$.

$\hfill \bullet$
\end{theorem}
\noindent Note that \cite{mishachev_backpropagation_2017} omits the vectorization operation on the right-hand side of eq. \eqref{eq:partial_gradient}. The theory of matrix differential calculus, however, requires the gradient of a matrix-variate real-valued function to be a column vector (cf.\secref{sec:appendix_1}). An exemplary application of the BP algorithm for the 3-layered neural network introduced in Example 1 is visualized in \autoref{fig:figure_3}B. 

The motivation for using the BP algorithm in lieu of a standard numerical differentiation algorithm for evaluating the partial gradients of training exemplar-specific cost functions is an immense reduction in the number of necessary computations: as evident from eqs. \eqref{eq:w_update} and \eqref{eq:c_gradient}, neural network gradient descent requires the evaluation of the training exemplar-specific cost functions' gradients on each iteration of the algorithm. Each of these gradients comprises the partial derivatives
\begin{equation}\label{eq:partial_derivatives}
\frac{\partial}{\partial w_{ij}^{l}} c_{xy}(W^1,...,W^k)  
\mbox{ for all } 
i = 1,...,n_l, j = 1,...,n_{l-1}+1, \mbox{ and } l = 1,...,k.
\end{equation}
A naive approach to the numerical evaluation of the gradient of a $c_{xy}$ is to approximate the derivatives in eq. \eqref{eq:partial_derivatives} by means of difference quotients of the form
\begin{equation}
\frac{\partial}{\partial w_{ij}^{l}} c_{xy}(W^1,..., W^k) 
\approx
\frac{c_{xy}(W^1,..., \tilde{W}^l,..., W^k) - c_{xy}(W^1,..., W^l, ...,W^k)}{\epsilon},
\end{equation}
where
\begin{equation}
\tilde{W}^l := W^l + 1_{ij}^{l}\epsilon ,
\end{equation}
$\epsilon > 0$ is a suitably chosen small parameter, and $1_{ij}^{l} \in \mathbb{R}^{n_l \times (n_{l-1}+1)}$ denotes a matrix of all $0$'s except for a $1$ at the location corresponding to the entry of $w_{ij}^{l}$ in $W^l$. Notably, for each iteration of the gradient descent and for all training exemplars, such an approach would require $1 + \sum_{l = 1}^k n_{l}({n_{l-1}}+1)$ evaluations of the training exemplar-specific cost function $c_{xy}$ and thus of the neural network $f$. In common nomenclature, the evaluation of $f$ for a given training input exemplar $x$ is referred to as a ``forward pass''. The key feature of the BP algorithm for the evaluation of $\nabla c_{xy}$ is that it reduces the necessary number of forward passes for evaluating $\nabla c_{xy}$ on a given iteration of the gradient descent algorithm from  $1 + \sum_{l = 1}^k n_{l}({n_{l-1}}+1)$ to $1$. To achieve this, the BP adds the ``backward pass'' defined in \autoref{thm:BP_algorithm}, which is of similar computational complexity as the forward pass. Thus, the BP algorithm reduces the number of necessary computational steps for evaluating $\nabla c_{xy}$ from  $1 + \sum_{l = 1}^k n_{l}({n_{l-1}}+1)$ to $2$. 

\section{Proof of the BP algorithm}\label{sec:proof}
We show the validity of the BP recursion by induction with respect to the number of layers $k$ of a neural network. To this end, we first validate the BP recursion directly for $k := 3$, i.e., for a  3-layered neural network (\hspace{-1mm}\secref{sec:base_case}). We then assume the validity of the BP recursion for some $k$ and show that it is also valid for a neural network with $k+1$ layers (\hspace{-1mm}\secref{sec:inductive_step}). 

\subsection{Base case}\label{sec:base_case} 
\noindent We consider the case of a 3-layered homogeneous neural network, i.e., a multivariate vector-valued function of the form (cf. \defref{def:neural_network})
\begin{equation}\label{eq:proof_f}
f : \mathbb{R}^{n_0} \to \mathbb{R}^{n_3}, 
x \mapsto f(x) := 
\Sigma^{3}(\Phi^3_{W^3}(\Sigma^{2}(\Phi^2_{W^2}(\Sigma^{1}(\Phi^1_{W^1}(x)))))) 
\end{equation}
with associated weight matrix-variate neural network function 
(cf. \defref{def:matrix_variate_neural_network_function})
\begin{equation}\label{eq:proof_fx}
f_x : 
       \mathbb{R}^{n_1 \times (n_0 + 1)} 
\times \mathbb{R}^{n_2 \times (n_1 + 1)} 
\times \mathbb{R}^{n_3 \times (n_2 + 1)}
\to 
\mathbb{R}^{n_3},\,
(W^1, W^2, W^3) \mapsto
f_x(W^1, W^2, W^3).
\end{equation}
The network's potential vector and activation vector definitions resulting from \defref{def:potential_functions} and \defref{def:neural_network} are listed in \autoref{tab:table_3}. We consider the training exemplar-specific cost function (cf. \defref{def:cost_functions}, eq. \eqref{eq:training_exemplar_cost_function})
\begin{multline}\label{eq:proof_cxy}
c_{xy} : 
	   \mathbb{R}^{n_1 \times (n_0+1)} 
\times \mathbb{R}^{n_2 \times (n_1+1)} 
\times \mathbb{R}^{n_3 \times (n_2+1)}
\to 
\mathbb{R}, 
\\
(W^1, W^2, W^3) \mapsto
c_{xy}(W^1,W^2,W^3)
:= c_y(f_x(W^1,W^2,W^3)) 
\end{multline}
and the output-specific loss function (cf. \defref{def:cost_functions}, eq. \eqref{eq:output_cost_function})
\begin{equation}
c_y : \mathbb{R}^{n_3} \to \mathbb{R}, a^3 \mapsto c_y(a^3).
\end{equation}
For $l = k = 3, l = k - 1 =  2$, and $l = k - 2 = 1$, we show below that the formal application of the BP algorithm as defined in \autoref{thm:BP_algorithm} yields an expression $\tilde{\nabla}_{W^l}c_{xy} (W^1, W^2, W^3)$ that is identical to the partial gradient $\nabla_{W^l}c_{xy} (W^1, W^2, W^3)$ as evaluated by matrix differential calculus. To this end, we note that with the results of \autoref{sec:appendix_3}, the Jacobian matrix of a weight matrix-variate potential function $\Phi_a$ at $W$ (cf. eq. \eqref{eq:Phi_a}) is given by 
\begin{equation}\label{eq:DPhi_a}
\mbox{D}\Phi_a(W) = (a^T \,\, 1) \otimes I_m,
\end{equation}
the Jacobian matrix of a potential function $\Phi_W$ at $a$ (cf. eq. \eqref{eq:Phi_W}) is given by 
\begin{equation}\label{eq:DPhi_W}
\mbox{D}\Phi_W(a) = W_\bullet,
\end{equation}
and the Jacobian matrix of a component-wise activation function $\Sigma$ at $z$ (cf. eq. \eqref{eq:Sigma}) is given by 
\begin{equation}\label{eq:DSigma}
\mbox{D}\Sigma(z) = \mbox{diag}(\sigma'(z_1), ..., \sigma'(z_n)).
\end{equation}

\begin{table}[t!]
\begin{center}
\renewcommand{\arraystretch}{1.5}
\begin{tabular}{lll}
  $z^1$ 
& $ := \Phi^1(W^1,x)$ 
& $ := W^1 \cdot (x^T \,\, 1)^T$ 
\\
  $a^1$ 
& $ := \Sigma^1(\Phi^1(W^1,x))$ 
& $ := \Sigma^1(W^1 \cdot (x^T \,\, 1)^T)$ 
\\
  $z^2$ 
& $ := \Phi^2(W^2,\Sigma^1(\Phi^1(W^1,x)))$ 
& $ := W^2\cdot \Sigma^1(W^1\cdot (x^T \,\, 1)^T)$ 
\\
  $a^2$ 
& $ := \Sigma^2(\Phi^2(W^2,\Sigma^1(\Phi^1(W^1,x))))$ 
& $ := \Sigma^2(W^2\cdot\Sigma^1(W^1\cdot (x^T \,\, 1)^T))$ 
\\
  $z^3$ 
& $ := \Phi^3(W^3,\Sigma^2(\Phi^2(W^2,\Sigma^1(\Phi^1(W^1,x)))))$ 
& $ := W^3\cdot\Sigma^2(W^2\cdot\Sigma^1(W^1\cdot (x^T \,\, 1)^T))$ 
\\
  $a^3$ 
& $ := \Sigma^3(\Phi^3(W^3,\Sigma^2(\Phi^2(W^2,\Sigma^1(\Phi^1(W^1,x))))))$ 
& $ := \Sigma^3(W^3\cdot\Sigma^2(W^2\cdot\Sigma^1(W^1\cdot (x^T \,\, 1)^T)))$ 
\end{tabular}
\end{center}
\caption{Potential and activation vector definitions for a 3-layered neural network.}\label{tab:table_3}
\end{table}

\subsubsection*{Evaluation of $\nabla_{W^3}c_{xy}(W^1,W^2,W^3)$}
\noindent We first note that the formal application of the BP algorithm for $l = k = 3$ yields (cf. eq. \eqref{eq:error})
\begin{align}
\begin{split}
\delta^3
& = ((W^4_\bullet)^T \cdot \delta^4)\circ \tilde{\Sigma}^3(z^3)   	\\ 
& = (1^T \cdot \nabla c_y(a^3)) \circ \tilde{\Sigma}^3(z^3) \\   
& = \nabla c_y(a^3) \circ \tilde{\Sigma}^3(z^3)  
\end{split}
\end{align}
and thus (cf. eq. \eqref{eq:partial_gradient})
\begin{align}\label{eq:bp_nabla_cxy_W3}
\begin{split}
\tilde{\nabla}_{W^3}c_{xy}(W^1,W^2,W^3) 
& = \vex(\delta^3\cdot (\begin{matrix} \Sigma^{2}(z^{2})^T & 1 \end{matrix})) \\	 
& = \vex((\nabla c_y(a^3) \circ \tilde{\Sigma}^{3}(z^3))\cdot(\Sigma^{2}(z^{2})^T \,\,\, 1)).
\end{split}
\end{align}
Our aim is to show that $\tilde{\nabla}_{W^3}c_{xy}(W^1,W^2,W^3)$ indeed corresponds to $\nabla_{W^3}c_{xy}(W^1,W^2,W^3)$. We have
\begin{align*}\label{eq:D_W_3}
\nabla_{W^3}c_{xy}(& W^1, W^2, W^3)		
\\
& = (\mbox{D}_{W^3}c_{xy}(W^1, W^2, W^3))^T
\\
& =
(\mbox{D}_{W^3}(c_y(f_x(W^1,W^2,W^3))))^T 
&&
{\footnotesize\mbox{(with \defref{def:cost_functions}, eq. \eqref{eq:training_exemplar_cost_function})}}
\\ 
& = 
(\mbox{D}(c_y(f_x^3(W_3))))^T 
&& {\footnotesize \mbox{(with \defref{def:matrix_variate_neural_network_function}, eq. \eqref{eq:f_x_l})}}
\\
& =
(\mbox{D}(c_y(\Sigma^3(\Phi^3(W^3,\Sigma^2(\Phi^2(W^2,\Sigma^1(\Phi^1(W^1,x)))))))))^T 		
&& {\footnotesize \mbox{(with \defref{def:matrix_variate_neural_network_function}, eq. \eqref{eq:f_x})}}
\\
& = 
(\mbox{D}c_y(\Sigma^3(\Phi^3(W^3,\Sigma^2(\Phi^2(W^2,\Sigma^1(\Phi^1(W^1,x))))))) 	\\
& \quad  
\cdot
\mbox{D}\Sigma^3(\Phi^3(W^3,\Sigma^2(\Phi^2(W^2, \Sigma^1(\Phi^1(W^1,x)))))) 		\\
& \quad
\cdot
\mbox{D}\,\Phi^3(W^3,\Sigma^2(\Phi^2(W^2,\Sigma^1(\Phi^1(W^1,x))))))^T
&& {\footnotesize \mbox{(with \autoref{thm:iterated_chain_rule_2})}}
\\
& = 
(\mbox{D}c_y(a^3)\cdot \mbox{D}\Sigma^3(z^3)\cdot\mbox{D}\Phi^3(W^3,a^2))^T 
&& {\footnotesize \mbox{(with \autoref{tab:table_3})}}
\\
& = 
(\mbox{D}c_y(a^3)\cdot \mbox{D}\Sigma^3(z^3)\cdot\mbox{D}\Phi^3_{a^2}(W^3))^T
&& {\footnotesize \mbox{(with \defref{def:potential_functions}, eq. \eqref{eq:Phi_a})}}
\\
& = \mbox{D}\Phi^3_{a^2}(W^3)^T \cdot \mbox{D}\Sigma^3(z^3)\cdot \mbox{D}c_y(a^3)^T 									
\\
& = (((a^2)^T \,\,\, 1) \otimes I_{n_3})^T \cdot (\nabla c_y(a^3) \circ \tilde{\Sigma}^{3}(z^3))			
&& {\footnotesize \mbox{(with eq. \eqref{eq:DPhi_a} and eq. \eqref{eq:hadamard_diagonal_matrix}})}
\\
& = (((a^2)^T \,\,\, 1)^T  \otimes I_{n_3}) \cdot (1 \otimes (\nabla c_y(a^3) \circ \tilde{\Sigma}^{3}(z^3))) 	
&& {\footnotesize \mbox{(with eq. } \eqref{eq:kron_transposition} \mbox{)}}
\\
& = (((a^2)^T \,\,\, 1)^T \cdot 1) \otimes (I_{n_3}\cdot(\nabla c_y(a^3) \circ \tilde{\Sigma}^{3}(z^3)))  					
&& {\footnotesize \mbox{(with eq. } \eqref{eq:kron_mixed} \mbox{)}}	
\\
& = ((a^2)^T \,\,\,1)^T \otimes (\nabla c_y(a^3) \circ \tilde{\Sigma}^{3}(z^3)) 	
&& 
\\
& = \vex((\nabla c_y(a^3) \circ \tilde{\Sigma}^{3}(z^3))\cdot ((a^2)^T \,\,\,1))								
&& {\footnotesize \mbox{(with eq. \eqref{eq:kron_vectorization} for } X := 1 \mbox{)}}	
\\
& = \vex((\nabla c_y(a^3) \circ \tilde{\Sigma}^{3}(z^3))\cdot (\Sigma^2(z^2)^T \,\,\, 1)) 
&& {\footnotesize \mbox{(with \autoref{tab:table_3})}},
\end{align*}
where we included explicit justifications for selected equalities which will also be essential in the remainder of the proof.

\subsubsection*{Evaluation of $\nabla_{W^2}c_{xy}(W^1,W^2,W^3)$}
\noindent We first note that the formal application of the BP algorithm for $l = k - 1 = 3 - 1 = 2$ yields
\begin{align}
\begin{split}
\delta^{2}  
& = ((W^3_\bullet)^T \cdot \delta^{3}) \circ \tilde{\Sigma}^{2}(z^2)										\\
& = ((W^3_\bullet)^T \cdot (\nabla c_y(a^3)\circ \tilde{\Sigma}^3(z^3))) \circ \tilde{\Sigma}^{2}(z^{2})
\end{split}
\end{align}
and thus with \eqref{eq:partial_gradient}
\begin{align}\label{eq:bp_nabla_cxy_W2}
\begin{split}
\tilde{\nabla}_{W^2}c_{xy}(W^1,W^2,W^3) 
& = \vex(\delta^{2}\cdot(\Sigma^{1}(z^{1})^T) 																			\\
& = \vex(((W^3_\bullet)^T\cdot(\nabla c_y(a^3)\circ\tilde{\Sigma}^3(z^3))\circ\tilde{\Sigma}^{2}(z^{2}))
     \cdot (\Sigma^1(z^1)^T \,\,\, 1)).
\end{split}
\end{align}
Our aim is to show that $\tilde{\nabla}_{W^2}c_{xy}(W^1,W^2,W^3)$ indeed corresponds to $\nabla_{W^2}c_{xy}(W^1,W^2,W^3)$. To this end, we have
\begin{align}\label{eq:D_W_2}
\begin{split}
\nabla_{W^2}c_{xy}(W^1, W^2 &, W^3)
\\																		
& = (\mbox{D}_{W^2} c_{xy}(W^1, W^2, W^3))^T														
\\
& = (\mbox{D}_{W^2}(c_y(f_x(W^1,W^2,W^3))))^T 							
\\ 
& = (\mbox{D}(c_y(f_x^2(W_2))))^T										
\\
& = (\mbox{D}(c_y(\Sigma^3(\Phi^3(W^3,\Sigma^2(\Phi^2(W^2,\Sigma^1(\Phi^1(W^1,x)))))))))^T  	
\\
& =	(\mbox{D}c_y(\Sigma^3(\Phi^3(W^3,\Sigma^2(\Phi^2(W^2,\Sigma^1(\Phi^1(W^1,x))))))) 	\\
	& \quad  
	\cdot
	\mbox{D}\Sigma^3(\Phi^3(W^3,\Sigma^2(\Phi^2(W^2, \Sigma^1(\Phi^1(W^1,x))))) 		\\
	& \quad
	\cdot
	\mbox{D}\Phi^3(W^3,\Sigma^2(\Phi^2(W^2,\Sigma^1(\Phi^1(W^1,x)))))					\\
	& \quad
	\cdot
	\mbox{D}\Sigma^2(\Phi^2(W^2,\Sigma^1(\Phi^1(W^1,x))))								\\
	& \quad
	\cdot
	\mbox{D}\Phi^2(W^2,\Sigma^1(\Phi^1(W^1,x))))^T														
\\
& = (
		\mbox{D}c_y(a^3)
		\cdot \mbox{D}\Sigma^3(z^3)
		\cdot \mbox{D}\Phi(W^3,a^2)
		\cdot \mbox{D}\Sigma^2(z^2) 
		\cdot \mbox{D}\Phi(W^2,a^1)
	)^T														
\\
& = (	\mbox{D}c_y(a^3)
		\cdot \mbox{D}\Sigma^3(z^3)
		\cdot \mbox{D}\Phi^3_{W^3}(a^2)
		\cdot \mbox{D}\Sigma^2(z^2)
		\cdot \mbox{D}\Phi^2_{a^1}(W^2)
	)^T
\\
& = \mbox{D}\Phi^2_{a^1}(W^2)^T
	\cdot \mbox{D}\Sigma^2(z^2)
	\cdot \mbox{D}\Phi^3_{W^3}(a^2)^T
	\cdot \mbox{D}\Sigma^3(z^3)
	\cdot \mbox{D}c_y(a^3)^T			
\\
& = \mbox{D}\Phi^2_{a^1}(W^2)^T
	\cdot \mbox{D}\Sigma^2(z^2)
	\cdot (W^3_\bullet)^T
	\cdot (\nabla c_y(a^3) \circ \tilde{\Sigma}^{3}(z^3)) 
\\
& = \mbox{D}\Phi^2_{a^1}(W^2)^T
	\cdot
	(
	\tilde{\Sigma}(z^2)
	\circ
	((W^3_\bullet)^T \cdot (\nabla c_y(a^3) \circ \tilde{\Sigma}^{3}(z^3)))
	)
\\
& = \mbox{D}\Phi^2_{a^1}(W^2)^T
	\cdot
	(
	(W^3_\bullet)^T \cdot (\nabla c_y(a^3) \circ \tilde{\Sigma}^{3}(z^3))
	\circ
	\tilde{\Sigma}(z^2)
	)
\\
& = (((a^1)^T \,\,\, 1) \otimes  I_{n_2})^T
	\cdot
	((W^3_\bullet)^T \cdot (\nabla c_y(a^3) \circ \tilde{\Sigma}^{3}(z^3))
	\circ
	\tilde{\Sigma}(z^2)
	)
\\
& = (((a^1)^T \,\,\, 1)^T \otimes  I_{n_2})
	\cdot
	(1 \otimes ((W^3_\bullet)^T \cdot (\nabla c_y(a^3) \circ \tilde{\Sigma}^{3}(z^3))
	\circ
	\tilde{\Sigma}(z^2)))
\\
& = (((a^1)^T \,\,\, 1)^T \cdot 1)
	\otimes
	(I_{n_2} \cdot((W^3_\bullet)^T \cdot (\nabla c_y(a^3) \circ \tilde{\Sigma}^{3}(z^3))
	\circ
	\tilde{\Sigma}(z^2))
\\
& = ((a^1)^T \,\,\, 1)^T \otimes
	((W^3_\bullet)^T \cdot (\nabla c_y(a^3) \circ \tilde{\Sigma}^{3}(z^3))\circ\tilde{\Sigma}(z^2)))
\\
& = \vex(((W^3_\bullet)^T \cdot (\nabla c_y(a^3) \circ \tilde{\Sigma}^{3}(z^3))\circ \tilde{\Sigma}(z^2))
    \cdot((a^1)^T \,\,\, 1)) 
\\
& = \vex(((W^3_\bullet)^T \cdot (\nabla c_y(a^3) \circ \tilde{\Sigma}^{3}(z^3))\circ \tilde{\Sigma}(z^2))
	\cdot (\Sigma^1(z^1)^T \,\,\, 1)). 
\end{split}
\end{align}

\subsubsection*{Evaluation of $\nabla_{W^1}c_{xy}(W^1,W^2,W^3)$}
\noindent We first note that the formal application of the BP algorithm for $l = k - 2 = 3 - 2 = 1$ yields
\begin{align}
\begin{split}
\delta^1 
& = ((W^2_\bullet)^T \cdot \delta^2)
\circ \tilde{\Sigma}^1(z^1) \\
& = ((W^2_\bullet)^T \cdot (((W^3_\bullet)^T \cdot (\nabla c_y(a^3)\circ \tilde{\Sigma}^3(z^3))) \circ \tilde{\Sigma}^{2}(z^{2})))
\circ \tilde{\Sigma}^1(z^1) \\
\end{split}
\end{align}
and thus
\begin{align}
\begin{split}
\tilde{\nabla}_{W^1}c_{xy}(W^1, & W^2, W^3) \\
& = \vex(\delta^1\cdot(\Sigma^{0}(z^0)^T \,\,\,  1))\\
& = \vex
	(
	(
	(
	(W^2_\bullet)^T \cdot
	(
	(
	(W^3_\bullet)^T 
	\cdot (\nabla c_y(a^3)
	\circ \tilde{\Sigma}^3(z^3)
	)
	)
	\circ 
	\tilde{\Sigma}^{2}(z^{2})
	)
	)
	\circ \tilde{\Sigma}^1(z^1)
	) 
	\cdot (x^T \,\,\, 1)
	).
\end{split}
\end{align}
Our aim is to show that $\tilde{\nabla}_{W^1}c_{xy}(W^1,W^2,W^3)$ indeed corresponds to $\nabla_{W^1}c_{xy}(W^1,W^2,W^3)$. To this end, we have
\begin{align*}\label{eq:D_W_1}
\begin{split}
\nabla_{W^1}c_{xy}(W^1 &,W^2, W^3)		\\																
& = (\mbox{D}_{W^1} c_{xy}(W^1, W^2, W^3))^T														
\\
& = (\mbox{D}_{W^1}(c_y(f_x(W^1,W^2,W^3))))^T 							
\\ 
& = (\mbox{D}(c_y(f_x^1(W_1))))^T										
\\
& = (\mbox{D}(c_y(\Sigma^3(\Phi^3(W^3,\Sigma^2(\Phi^2(W^2,\Sigma^1(\Phi^1(W^1,x)))))))))^T  	
\\
& =	(\mbox{D}c_y(\Sigma^3(\Phi^3(W^3,\Sigma^2(\Phi^2(W^2,\Sigma^1(\Phi^1(W^1,x)))))))) 	\\
	& \quad  
	\cdot
	\mbox{D}\Sigma^3(\Phi^3(W^3,\Sigma^2(\Phi^2(W^2, \Sigma^1(\Phi^1(W^1,x))))) 		\\
	& \quad
	\cdot
	\mbox{D}\Phi^3(W^3,\Sigma^2(\Phi^2(W^2,\Sigma^1(\Phi^1(W^1,x)))))					\\
	& \quad
	\cdot
	\mbox{D}\Sigma^2(\Phi^2(W^2,\Sigma^1(\Phi^1(W^1,x))))								\\
	& \quad
	\cdot
	\mbox{D}\Phi^2(W^2,\Sigma^1(\Phi^1(W^1,x)))											\\						
	& \quad
	\cdot
	\mbox{D}\Sigma^1(\Phi^1(W^1,x))														\\
	& \quad
	\cdot
	\mbox{D}\Phi^1(W^1,x))^T															\\
& = (
	\mbox{D}c_y(a^3)
	\cdot \mbox{D}\Sigma^3(z^3)
	\cdot \mbox{D}\Phi(W^3,a^2)
	\cdot \mbox{D}\Sigma^2(z^2) 
	\cdot \mbox{D}\Phi^1(W^2,a^1)
	\cdot \mbox{D}\Sigma^1(z^1)							
	\cdot \mbox{D}\Phi^1(W^1,x)							
	)^T														
\\
& = (
	\mbox{D}c_y(a^3)
	\cdot \mbox{D}\Sigma^3(z^3)
	\cdot \mbox{D}\Phi_{W^3}^3(a^2)
	\cdot \mbox{D}\Sigma^2(z^2) 
	\cdot \mbox{D}\Phi^2_{W^2}(a^1)
	\cdot \mbox{D}\Sigma^1(z^1)							
	\cdot \mbox{D}\Phi^1_x(W^1)							
	)^T	
\\
& = \mbox{D}\Phi^1_x(W^1)^T	
	\cdot \mbox{D}\Sigma^1(z^1)	
	\cdot \mbox{D}\Phi_{W^2}^2(a^1)^T	
	\cdot \mbox{D}\Sigma^2(z^2) 
  	\cdot \mbox{D}\Phi_{W^3}^3(a^2)^T	
	\cdot \mbox{D}\Sigma^3(z^3)	
	\cdot \mbox{D}c_y(a^3)^T									
\\
& = \mbox{D}\Phi^1_x(W^1)^T	
	\cdot	\mbox{D}\Sigma^1(z^1)	
	\cdot \mbox{D}\Phi_{W^2}^2(a^1)^T	
	\cdot 	(
			(W^3_\bullet)^T \cdot 
			(\nabla c_y(a^3) \circ \tilde{\Sigma}^{3}(z^3))
			\circ
			\tilde{\Sigma}(z^2)
			)
\\
& = \mbox{D}\Phi^1_x(W^1)^T	
  	\cdot	\mbox{D}\Sigma^1(z^1)	
	\cdot (W^2_\bullet)^T	
	\cdot 	(
	  		(W^3_\bullet)^T \cdot 
			(\nabla c_y(a^3) \circ \tilde{\Sigma}^{3}(z^3))
			\circ
			\tilde{\Sigma}(z^2)
			)
\\
& =	\mbox{D}\Phi^1_x(W^1)^T	
	\cdot		  
	(\tilde{\Sigma}^1(z^1) \circ
	((W^2_\bullet)^T	
	\cdot 	(
			(W^3_\bullet)^T \cdot 
			(\nabla c_y(a^3) \circ \tilde{\Sigma}^{3}(z^3))
			\circ
			\tilde{\Sigma}(z^2)
			)
  	)
  	)
\\
& =   \mbox{D}\Phi^1_x(W^1)^T	
	  \cdot	  
	  (
	  ((W^2_\bullet)^T	
	  \cdot (
	  		(W^3_\bullet)^T \cdot 
			(\nabla c_y(a^3) \circ \tilde{\Sigma}^{3}(z^3))
			\circ
			\tilde{\Sigma}(z^2)
			)
	  )
	  \circ \tilde{\Sigma}^1(z^1) 
	  )
\\
& =   ((x^T \,\,\, 1) \otimes I_{n_1})^T
	  \cdot	  
	  (
	  (
	  (W^2_\bullet)^T	
	  \cdot (
	  		(W^3_\bullet)^T \cdot 
			(\nabla c_y(a^3) \circ \tilde{\Sigma}^{3}(z^3))
			\circ
			\tilde{\Sigma}(z^2)
			)
	  )
	  \circ \tilde{\Sigma}^1(z^1) 
	  )
\\
& =   ((x^T \,\,\, 1)^T \otimes I_{n_1})
	  \cdot
	  (1 \otimes
	  (
	  (W^2_\bullet)^T	
	  \cdot (
	  		(W^3_\bullet)^T \cdot 
			(\nabla c_y(a^3) \circ \tilde{\Sigma}^{3}(z^3))
			\circ
			\tilde{\Sigma}(z^2)
			)
	  )
	  \circ \tilde{\Sigma}^1(z^1) 
	  )
\\
& =   ((x^T \,\,\, 1)^T \cdot 1)
	  \otimes
	  (I_{n_1} \cdot
	  ((W^2_\bullet)^T	
	  \cdot (
	  		(W^3_\bullet)^T \cdot 
			(\nabla c_y(a^3) \circ \tilde{\Sigma}^{3}(z^3))
			\circ
			\tilde{\Sigma}(z^2)
			)
	  )
	  \circ \tilde{\Sigma}^1(z^1) 
	  )
\\
& =   (x^T \,\,\, 1)^T
	  \otimes
	  (
	  ((W^2_\bullet)^T	
	  \cdot (
	  		(W^3_\bullet)^T \cdot 
			(\nabla c_y(a^3) \circ \tilde{\Sigma}^{3}(z^3))
			\circ
			\tilde{\Sigma}(z^2)
			)
	  )
	  \circ \tilde{\Sigma}^1(z^1) 
	  )
\\
& =   \vex
	  (
	  ((W^2_\bullet)^T	
	  \cdot (
	  		(W^3_\bullet)^T \cdot 
			(\nabla c_y(a^3) \circ \tilde{\Sigma}^{3}(z^3))
			\circ
			\tilde{\Sigma}(z^2)
			)
	  \circ \tilde{\Sigma}^1(z^1) 
	  )
	  \cdot(x^T \,\,\, 1)
	  ).
\end{split}
\end{align*}
This completes the base case validation.

\subsection{Inductive step}\label{sec:inductive_step}
\noindent We assume that the induction hypothesis holds for $k = \bar{k}$ (e.g., for $k = 3$). More explicitly, we have the following \textit{induction hypothesis (H)}, which we assume to hold:
\begin{enumerate}
\item[(H)] For a $\bar{k}$-layered neural network, it holds that for $l = \bar{k}, \bar{k} - 1, ...,  \bar{k} - (\bar{k}-1)$
\begin{equation}
\nabla_{W^l} 
c_{xy}(W^1, ..., W^{\bar{k}}) 
:= 	\vex
	(
	(
	(W^{l+1}_\bullet)^T \cdot \delta^{l+1}
	) 
	\circ \tilde{\Sigma}^{l}(z^l)
	)
	\cdot  
	(\Sigma^{l-1}(z^{l-1})^T \,\,\, 1)
	),
\end{equation}
with $W^{\bar{k}+1} := 1$, $\delta^{\bar{k}+1} := \nabla c_y(a^{\bar{k}})$, and $\Sigma^0(z^0):= x$.
\end{enumerate}
We claim, that if (H) is true, the following \textit{induction claim (C)} for $k = \bar{k} + 1$ is also true:
\begin{enumerate}
\item[(C)] For a $\bar{k}+1$-layered neural network, it holds that for $l = \bar{k}+1, (\bar{k}+1)-1, ..., (\bar{k}+1) - ((\bar{k}+1)-1)$
\begin{equation}
\nabla_{W^l} 
c_{xy}(W^1, ..., W^{\bar{k}+1}) 
:= 	\vex
	(	
	(
	(
	(W^{l+1}_\bullet)^T \cdot \delta^{l+1}
	) 
	\circ \tilde{\Sigma}^{l}(z^l) 
	)
	\cdot 
	(\Sigma^{l-1}(z^{l-1})^T \,\,\, 1)
	),
\end{equation}
with $W^{(\bar{k}+1)+1} := 1$, $\delta^{(\bar{k}+1)+1} := \nabla c_y(a^{\bar{k}+1})$, and $\Sigma^0(z^0) := x$.
\end{enumerate}
To see this, we set $\bar{\bar{k}} = \bar{k} + 1$. Then (C) corresponds to the following \textit{statement (S)}:
\begin{enumerate}
\item[(S)] For a $\bar{\bar{k}}$-layered neural network, it holds that for $l = \bar{\bar{k}}, \bar{\bar{k}}-1, ..., \bar{\bar{k}} - (\bar{\bar{k}}-1)$
\begin{equation}
\nabla_{W^l} 
c_{xy}(W^1, ..., W^{\bar{\bar{k}}}) 
:= 	
	\vex
	(
	(	
	((W^{l+1}_\bullet)^T \cdot \delta^{l+1}) 
	\circ \tilde{\Sigma}^{l}(z^l)
	)  
	\cdot (\Sigma^{l-1}(z^{l-1})^T \,\,\, 1)
	)
\end{equation}
with $W^{\bar{\bar{k}}+1} := 1$, $\delta^{\bar{\bar{k}}+1} := \nabla c_y(a^{\bar{\bar{k}}})$, and $\Sigma^0(z^0) := x$.
\end{enumerate}
(S) is identical to (H) except for the (arbitrary) denotation of the index $\bar{\bar{k}}$ which denotes the number of neural network layers. Hence, the induction claim holds under the assumption that the induction hypothesis holds for at least one $\bar{k}$. We have previously seen that it holds for $\bar{k} = 3$. Hence, the induction claim is justified.

$\hfill \Box$

\section{Exemplary application}\label{sec:exemplary_application}
\begin{figure}[t!]
\begin{center}
\includegraphics[width = \textwidth]{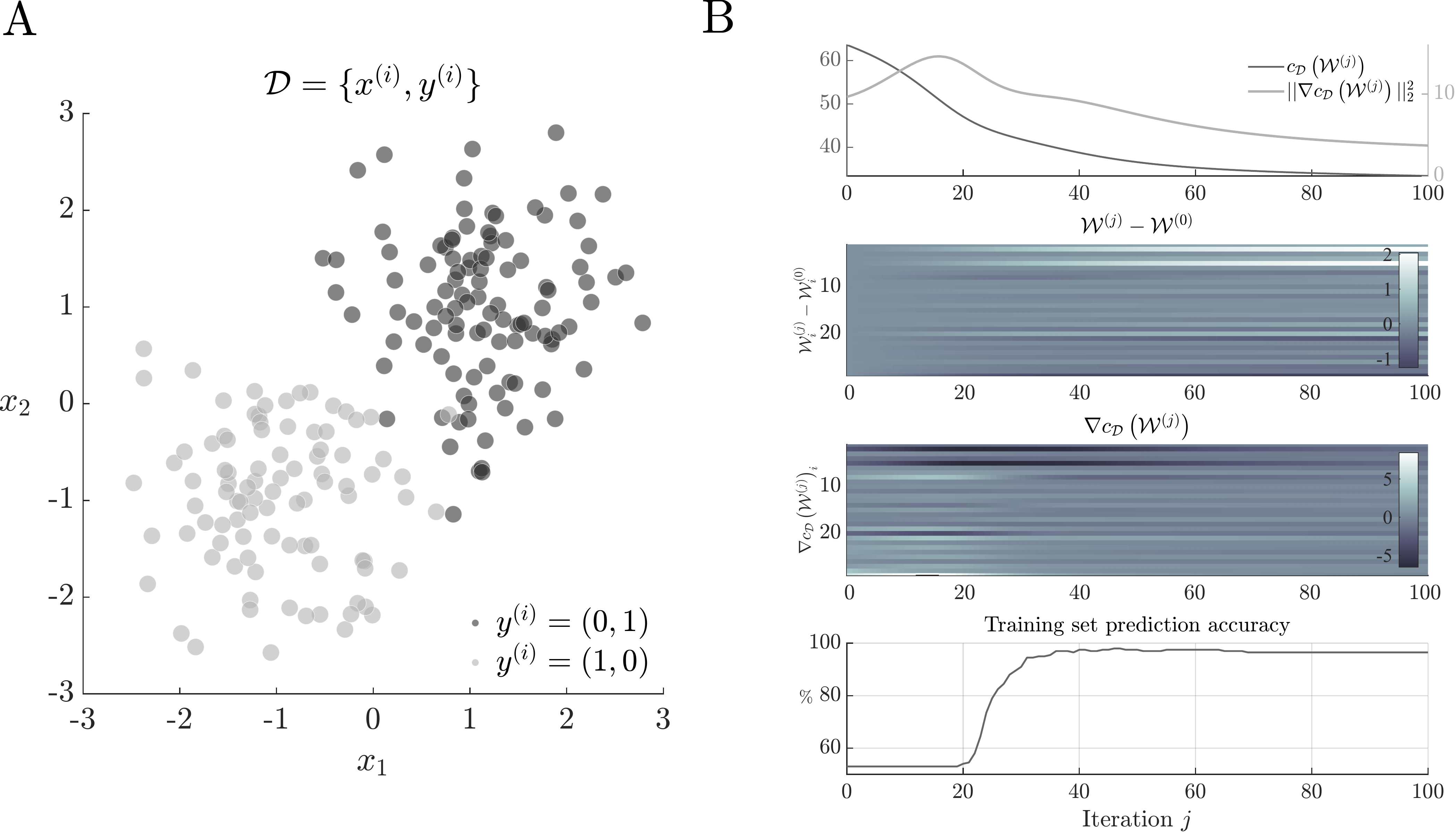}
\end{center}
\caption{Exemplary application. (A) Training data set sampled from a probabilistic model $p(x,y)$ as specified in eqs. \eqref{eq:p_marginal} and \eqref{eq:p_conditional}. Neural network training on the training set visualized in (A) for a 3-layered neural network as specified in \nameref{sec:example_1}. (B) The first panel depicts the evolution of the neural network's additive cost function and its associated gradient norm over iterations $j = 0,1,...,100$ of the gradient descent algorithm specified in \defref{def:gradient_descent} for $\alpha := 1$. The second panel depicts the evolution of the vector sequence $\mathcal{W}^{(j)} - \mathcal{W}^{(0)}$. The third panel visualizes the concomitant evolution of the additive cost function gradient vector. The fourth panel depicts the training set prediction accuracy of the neural network as a function of iterations.} \label{fig:figure_4}
\end{figure}

\autoref{fig:figure_4} visualizes an exemplary application of the BP algorithm for the gradient descent-based learning of the parameters of a 3-layered neural network. Specifically, a training data set $\mathcal{D} = \lbrace x^{(i)},y^{(i)} \rbrace_{i=1}^n$ comprising $n = 200$ target vectors $y^{(i)} \in \lbrace (0,1)^T, (1,0)^T\rbrace$  and feature vectors $x^{(i)} \in \mathbb{R}^2$ was first sampled from a probabilistic model of the form
\begin{equation}
p(x,y) = p(x|y)p(y),  
\end{equation}
where 
\begin{equation}\label{eq:p_marginal}
p(y = (1,0)^T) := 0.5 
\mbox{ and } 
p(y = (0,1)^T) := 0.5 
\end{equation}
denote the marginal probability mass functions of the target vector and 
\begin{equation}\label{eq:p_conditional}
p(x|y) = y_1N(x; \mu_0, \Sigma) + y_2N(x; \mu_1, \Sigma)
\mbox{ with }
\mu_0 := (-1,-1)^T, \mu_1 := (1,1)^T \mbox{, and }\Sigma := 0.5I_2
\end{equation}
denotes the conditional probability density function of the feature vector. The feature vectors $x^{(i)}, i = 1,...,n$ of the training data set are visualized in \autoref{fig:figure_4}A.

The weight matrices of a 3-layered neural network ($k = 3$) with a two-dimensional input layer ($n_0 = 2$), two three-dimensional hidden layers ($n_1 = 3$, $n_2 = 3$), and a two-dimensional output layer ($n_3 = 2$) (cf. \nameref{sec:example_1}) were initialized by sampling each weight value from a standard normal distribution, i.e.,
\begin{equation}
w_{ij}^l \sim N(0,1) \mbox{ for } i = 1,...,n_{l}, j = 1,...,n_{l-1}, l = 1,2,3.
\end{equation}
A gradient descent algorithm as described in \defref{def:gradient_descent} and with a learning rate parameter $\alpha := 1$ was then used to minimize the neural network's additive cost function $c_{\mathcal{D}}$. The first panel of \autoref{fig:figure_4}B depicts the evolution of the additive cost function values $c_\mathcal{D}(\mathcal{W}^{(j)})$ and the Euclidean norm of its gradient $||\nabla c_\mathcal{D}(\mathcal{W}^{(j)})||_2^2$ for $j = 0,1,...,100$ iterations of the gradient descent algorithm. The additive cost function decreases monotonically as expected. Similarly, the norm of its gradient decreases monotonically from iteration 40 onwards. The second panel of \autoref{fig:figure_4}B shows the concominant evolution of the neural network's weight vectors $\mathcal{W}^{(j)}$ relative to their initial value $\mathcal{W}^{(0)}$, i.e., the vector sequence $\mathcal{W}^{(j)} - \mathcal{W}^{(0)}$. As the additive cost function and the gradient norm start to level off from iteration 40 onwards, so does the evolution of this sequence. The third panel of \autoref{fig:figure_4}B visualizes the evolution of the additive cost function gradient $\nabla c_{\mathcal{D}}(\mathcal{W}^{(j)})$. The most prominent changes of the additive cost function gradient occur between iterations 20 and 40.  Finally, the fourth panel of \autoref{fig:figure_4}B depicts the training set prediction accuracy of the neural network, i.e., the average number of correctly classified feature vectors. The training set prediction accuracy remains at the chance level of 0.53 for the first 20 iterations upon which it steadily increases to a maximum value of 0.98 that is reached from iteration 40 onwards. In sum, the exemplary application discussed here demonstrates the ability of the BP algorithm introduced in \autoref{thm:BP_algorithm} to serve as a basis for gradient descent-based neural network training.

\newpage
\subsection{Conclusions}
In sum, we have provided a formal grounding of a core component of contemporary deep learning that we hope may not only serve as a didactic resource in the training of aspiring data scientists, but may also inspire the exploration of novel matrix analysis-based approaches in neural network training.

\begin{small}
\subsection*{Declarations}
\noindent\textbf{Conflicts of interest/Competing interests} Non declared.
\vspace{2mm}

\noindent\textbf{Code availability} All custom Matlab code implementing the simulations and analyses is available from the Open Science Framework at \url{https://osf.io/7awpj/}.
\vspace{2mm}

\noindent\textbf{Author contributions}
D.O.: conceptualization, formal analysis, investigation, methodology, project administration, 
resources, software, validation, visualization, writing - original draft, 
writing - review \& editing. F.U.: conceptualization, formal analysis, writing - review \& editing.
\end{small}

\bibliographystyle{apalike}
\bibliography{Referenzen}

\begin{thebibliography}{}

\bibitem[Alpaydin, 2014]{alpaydin_introduction_2014}
Alpaydin, E. (2014).
\newblock {\em Introduction to Machine Learning}.

\bibitem[Bishop, 2006]{bishop_pattern_2006}
Bishop, C.~M. (2006).
\newblock {\em Pattern Recognition and Machine Learning}.
\newblock Information Science and Statistics. {Springer}, {New York}.

\bibitem[Deisenroth et~al., 2020]{deisenroth_mathematics_2020}
Deisenroth, M.~P., Faisal, A.~A., and Ong, C.~S. (2020).
\newblock {\em Mathematics for {{Machine Learning}}}.
\newblock {Cambridge University Press}, first edition.

\bibitem[Duda et~al., 2001]{duda_pattern_2001}
Duda, R., Hart, P., and Stork, D. (2001).
\newblock {\em Pattern Classification}.
\newblock {Wiley}.

\bibitem[Goodfellow et~al., 2017]{goodfellow_deep_2017}
Goodfellow, I., Bengio, Y., and Courville, A. (2017).
\newblock {\em {Deep Learning}}.
\newblock {The Mit Press}, {Cambridge, Massachusetts}.

\bibitem[Haykin, 1998]{haykin_neural_1998}
Haykin, S.~S. (1998).
\newblock {\em Neural {{Networks}} - {{A Comprehensive Foundation}}}.
\newblock {Prentice Hall}, second edition.

\bibitem[Horn and Johnson, 1994]{horn_topics_1994}
Horn, R. and Johnson, C. (1994).
\newblock {\em Topics in Matrix Analysis}.
\newblock {Cambridge University Press}.

\bibitem[LeCun et~al., 2015]{lecun_deep_2015}
LeCun, Y., Bengio, Y., and Hinton, G. (2015).
\newblock Deep learning.
\newblock {\em Nature}, 521(7553):436--444.

\bibitem[Magnus and Neudecker, 1989]{magnus_matrix_1989}
Magnus, J.~R. and Neudecker, H. (1989).
\newblock Matrix {{Differential Calculus}} with {{Applications}} in
  {{Statistics}} and {{Econometrics}}.
\newblock {\em Journal of the American Statistical Association}, 84(408):1103.

\bibitem[Mishachev, 2017]{mishachev_backpropagation_2017}
Mishachev, N.~M. (2017).
\newblock Backpropagation in matrix notation.
\newblock {\em arXiv:1707.02746}.

\bibitem[Nielsen, 2015]{nielsen_neural_2015}
Nielsen, M. (2015).
\newblock {\em Neural {{Networks}} and {{Deep Learning}}}.
\newblock {Determination Press}.

\bibitem[Rumelhart et~al., 1986]{rumelhart_learning_1986a}
Rumelhart, D.~E., Hinton, G.~E., and Williams, R.~J. (1986).
\newblock Learning representations by back-propagating errors.
\newblock {\em Nature}, (323):533--536.

\bibitem[Schmidhuber, 2015]{schmidhuber_deep_2015}
Schmidhuber, J. (2015).
\newblock Deep learning in neural networks: {{An}} overview.
\newblock {\em Neural Networks}, 61:85--117.

\end{thebibliography}

\newpage
\renewcommand{\thesection}{Appendix \arabic{section}}
\renewcommand{\theequation}{A\arabic{section}.\arabic{equation}}
\setcounter{section}{0}

\section{Essentials of matrix differential calculus}\label{sec:appendix_1}
\setcounter{equation}{0}
\subsection*{Multivariate vector-valued functions}
Let
\begin{equation}
f : \mathbb{R}^{n} \to \mathbb{R}^{m}, x \mapsto f(x)
\end{equation}
denote a multivariate vector-valued function with multivariate real-valued component functions $f_i : \mathbb{R}^n \to \mathbb{R}, i = 1,...,m$. Then the matrix 
\begin{equation}\label{eq:jacobian_f}
\mbox{D}f(x) := 
\left(\frac{\partial}{\partial x_j} f_i(x)\right)_{1\le i\le m, 1 \le j \le n} 
\in \mathbb{R}^{m \times n},
\end{equation}
where $\frac{\partial}{\partial x_j}f_i(x)$ denotes the partial derivative of the $i$th component function with respect to $x_j$ is called the \textit{Jacobian matrix of $f$ at $x$}. The transpose of the Jacobian matrix is called the \textit{gradient of $f$ at $x$} and is denoted by
\begin{equation}\label{eq:gradient_f}
\nabla f(x) := (\mbox{D}f(x))^T \in \mathbb{R}^{n \times m}.
\end{equation}
Note that for a multivariate real-valued function $f:\mathbb{R}^n \to \mathbb{R}$, the Jacobian $\mbox{D}f(x) \in \mathbb{R}^{1 \times n}$ is a row vector and the gradient $\nabla f(x) \in \mathbb{R}^{n}$ is a column vector, as familiar from the theory of optimization. \citet[][Chapter 5.12]{magnus_matrix_1989} prove the following theorem.

\begin{theorem}[Chain rule for multivariate vector valued functions]\label{thm:chain_rule_1}
Let $S_0 \subset \mathbb{R}^{n_0}$ and assume that $f_1 : S_0 \to \mathbb{R}^{n_1}$ is differentiable at an interior point $x_0$ of $S_0$. Let $S_1\subset\mathbb{R}^{n_1}$ such that $f_1(x)\in S_1$ for all $x \in S_0$ and assume that $f_2 : \mathbb{R}^{n_1} \to \mathbb{R}^{n_2}$ is differentiable at an interior point $x_1 = f_1(x_0)$ of $S_1$. Then the composite function
\begin{equation}
f_2 \circ f_1 : \mathbb{R}^{n_0} \to \mathbb{R}^{n_2}, x \mapsto (f_2 \circ f_1)(x) := f_2(f_1(x))
\end{equation}
is differentiable at $x_0$ and
\begin{equation}
\mbox{D}(f_2 \circ f_1)(x_0) = (\mbox{D}f_2(x_1))(\mbox{D}f(x_0)).
\end{equation}
\begin{tiny}$\hfill\blacksquare$\end{tiny}
\end{theorem}
\noindent Note that $\mbox{D}f_1(x_0) \in \mathbb{R}^{n_1 \times n_0}$, $\mbox{D}f_2(x_1) \in \mathbb{R}^{n_2 \times n_1}$, and $\mbox{D}(f_2 \circ f_1)(x_0) \in \mathbb{R}^{n_2 \times n_0}$. The following generalization to more than two concatenated multivariate vector-valued functions follows immediately by induction.

\begin{theorem}[Iterated chain rule for multivariate vector-valued functions]\label{thm:iterated_chain_rule_1}
For $i = 0,1,...,k-1$, let $S_i \subset \mathbb{R}^{n_i}$. For $i = 1,2,...,k$, assume that $f_i : S_{i-1} \to \mathbb{R}^{n_i}$ is differentiable at an interior point $x_{i-1}$ of $S_{i-1}$ and that $f_i(x_{i-1}) \in S_i$ for all $x_{i-1} \in S_{i-1}$. Then the composite function
\begin{equation}
f_k \circ f_{k-1} \circ \cdots \circ f_1 
: S_0 \to \mathbb{R}^{n_k}, 
x \mapsto (f_k \circ f_{k-1} \circ \cdots \circ f_1)(x) 
:= f_k(f_{k-1}(\cdots (f_1(x))))
\end{equation}
is differentiable at $x_0$ and
\begin{equation}
\mbox{D}(f_k \circ f_{k-1} \circ \cdots \circ f_1)(x_0) 
= 
(\mbox{D}f_k(x_{k-1}))
(\mbox{D}f_{k-1}(x_{k-2}))
\cdots
(\mbox{D}f_1(x_0)).
\end{equation}
\begin{tiny}$\hfill\blacksquare$\end{tiny}
\end{theorem}

\subsection*{Matrix-variate matrix-valued functions}
To generalize the concepts introduced above to the case of matrix-variate matrix-valued functions, we first give the following definition.

\begin{definition}[Vectorization]
The \textit{vectorization} of a matrix $A \in \mathbb{R}^{m \times n}$ is defined as
\begin{equation}
\vex : \mathbb{R}^{m \times n} \to \mathbb{R}^{mn}, A \mapsto \vex(A)
:= (a_{11},...,a_{m1}, a_{12}, ...,a_{m2}, ..., a_{1n}, ...,a_{mn})^T.
\end{equation}
$\hfill \bullet$
\end{definition}
\noindent In words, $\vex(A)$ is obtained by stacking the columns of $A$ from left to right on top of each other. Let then 
\begin{equation}
F : \mathbb{R}^{n\times q} \to \mathbb{R}^{m \times p}, X \mapsto F(X)
\end{equation}
denote a matrix-variate matrix-valued function. \citet[][Chapter 5.15]{magnus_matrix_1989} observe that the calculus properties of matrix-variate functions follow immediately from the corresponding properties of multivariate functions, because instead of the matrix-variate matrix-valued function $F$ one may consider the \textit{equivalent multivariate vector-valued function} defined by
\begin{equation}
f : \mathbb{R}^{nq} \to \mathbb{R}^{mp}, \vex(X) \mapsto f(\vex(X)) := \vex(F(X)).
\end{equation}
The \textit{Jacobian matrix of $F$ at $X$} is then defined as
\begin{equation}\label{eq:jacobian_F}
\mbox{D}F(X) 
:= \mbox{D}f(\vex(X)) 
= \left(\frac{\partial}{\partial (\vex(X))_j} (\vex(F))_i(X)\right)_{1\le i\le mp, 1 \le j \le nq} 
\in \mathbb{R}^{mp \times nq},
\end{equation}
where $\partial/\partial (\vex(X))_j(\vex(F))_i(X)$ is the partial derivative of the $i$th component function of $\vex(F(X))$ with respect to the $j$th element of $\vex(X)$ evaluated at $X$. In other words, the Jacobian matrix of the matrix-variate matrix-valued function $F$ is defined as the Jacobian matrix of its equivalent multivariate vector-valued function $f$. The transpose of the Jacobian matrix of $F$ at $X$ is called the \textit{gradient of $F$ at $X$} and is denoted by
\begin{equation}\label{eq:gradient_F}
\nabla F(X) := (\mbox{D}F(X))^T \in \mathbb{R}^{nq \times mp}.
\end{equation}
Based on the definition of the Jacobian matrix of a matrix-variate matrix-valued function and \autoref{thm:chain_rule_1}, the following chain rule for the evaluation of the Jacobian matrix of concatenated matrix-variate matrix-valued functions then follows immediately \citep[][Chapter 5.15, Theorem 12]{magnus_matrix_1989}.

\begin{theorem}[Chain rule for matrix-variate matrix-valued functions]\label{thm:matrix_chain_rule}
Let $S_0\subset \mathbb{R}^{n_0\times q_0}$ and assume that $F_1 : S_0 \to \mathbb{R}^{n_1\times q_1}$ is differentiable at an interior point $X_0$ of $S_0$. Let $S_1 \subset\mathbb{R}^{n_1 \times q_1}$ such that $F_1(X)\in S_1$ for all $X \in S_0$ and assume that $F_2 : S_1 \to \mathbb{R}^{n_2 \times q_2}$ is differentiable at an interior point $X_1 = F_1(X_0)$ of $S_1$. Then the composite function
\begin{equation}
(F_2 \circ F_1) : \mathbb{R}^{n_0 \times q_0} \to \mathbb{R}^{n_2 \times q_2}, 
X \mapsto (F_2 \circ F_1)(X) := F_2(F_1(X))
\end{equation}
is differentiable at $X_0$ and
\begin{equation}
\mbox{D}(F_2 \circ F_1)(X_0) = (\mbox{D}F_2(X_1))(\mbox{D}F_1(X_0)).
\end{equation}
\begin{tiny}$\hfill\blacksquare$\end{tiny}
\end{theorem}
\noindent Note that $\mbox{D}F_1(X_0) \in \mathbb{R}^{n_1q_1\times n_0q_0}$, $\mbox{D}F_2(X_1) \in \mathbb{R}^{n_2q_2 \times n_1q_1}$, and  $\mbox{D}(F_2 \circ F_1)(X_0)) \in \mathbb{R}^{n_2q_2 \times n_0q_0}$. The following generalization to more than two concatenated matrix-variate matrix-valued functions then follows immediately by induction.
\begin{theorem}[Iterated chain rule for matrix-variate matrix-valued functions]\label{thm:iterated_chain_rule_2}
For $i = 0,1,...,k-1$, let $S_i \subset \mathbb{R}^{n_i \times q_i}$. For $i = 1,2,...,k$, assume that $F_i : S_{i-1} \to \mathbb{R}^{m_i \times p_i}$ is differentiable at an interior point $X_{i-1}$ of $S_{i-1}$ and that $F_i(X_{i-1}) \in S_i$ for all $X_{i-1} \in S_{i-1}$. Then the composite function
\begin{equation}
F_k \circ F_{k-1} \circ \cdots \circ F_1 
: S_0 \to \mathbb{R}^{n_k}, 
X \mapsto (F_k \circ F_{k-1} \circ \cdots \circ F_1)(x_0) 
:= F_k(F_{k-1}(\cdots (F_1(X))))
\end{equation}
is differentiable at $X_0$ and
\begin{equation}
\mbox{D}(F_k \circ F_{k-1} \circ \cdots \circ F_1)(X_0) 
= 
(\mbox{D}F_k(X_{k-1}))
(\mbox{D}F_{k-1}(X_{k-2}))
\cdots
(\mbox{D}F_1(X_0)).
\end{equation}
\begin{tiny}$\hfill\blacksquare$\end{tiny}
\end{theorem}

\subsection*{Multi-matrix-variate real-valued functions}
We refer to a function of the form
\begin{equation}
F : \mathbb{R}^{n_1 \times q_1} \times \cdots \times \mathbb{R}^{n_k \times q_k} \to \mathbb{R},
(X_1,...,X_k) \mapsto F(X_1, ..., X_k)
\end{equation}
as a \textit{multi-matrix-variate real-valued function}. The equivalent multivariate real-valued function of a multi-matrix-variate real-valued function is given by
\begin{equation}
\small
f : \mathbb{R}^{\sum_{l = 1}^k n_lq_l} \to \mathbb{R},
(\vex(X_1)^T, ..., \vex(X_k)^T)^T \mapsto  
f(\vex(X_1)^T, ..., \vex(X_k)^T)^T 
:= F(X_1,...,X_k).
\end{equation}
In analogy to the Jacobian matrix of a matrix-variate matrix-valued function (cf. eq. \eqref{eq:jacobian_F}), the Jacobian matrix of $F$ at $(X_1,...,X_k)$ is defined as the Jacobian matrix of its equivalent multivariate real-valued function,
\begin{equation}
\mbox{D}F(X_1,..., X_k) := \mbox{D}f((\vex(X_1)^T, ..., \vex(X_k)^T)^T )
\in \mathbb{R}^{1 \times \sum_{l = 1}^k n_lq_l}.
\end{equation}
The transpose of the Jacobian matrix of $F$ at $(X_1,...,X_k)$ is called the gradient of $F$ at $(X_1,...,X_k)$ and is denoted by
\begin{equation}
\nabla F(X_1,...,X_k) := (\mbox{D}F(X_1,...,X_k))^T \in \mathbb{R}^{\sum_{l = 1}^k n_lq_l}.
\end{equation}
Finally, for $l = 1,..., k$, we define the \textit{partial Jacobian matrix with respect to $X_l$ of $F$ at $(X_1,..., X_k)$} as
\begin{equation}
\mbox{D}_{X_l}F(X_1,...,X_k)
= 
\left(
\frac{\partial}{\partial(\vex(X_l))_j}
F(X_1,...,X_k)
\right)_{1 \le j \le n_lq_l}
\in \mathbb{R}^{1 \times n_lq_l}
\end{equation}
and the \textit{partial gradient with respect to $X_l$ of $F$ at $(X_1,..., X_k)$} as
\begin{equation}
\nabla_{X_l}F(X_1,...,X_k) := (\mbox{D}_{X_l}F(X_1,...,X_k) )^T \in \mathbb{R}^{n_lq_l}.
\end{equation}
Note that if for constant $X_\ell \in \mathbb{R}^{n_\ell \times n_{\ell-1}}, \ell = 1,...,k, \ell \neq l$ a function $F^l$ is defined by
\begin{equation}
F_l : \mathbb{R}^{n_l \times n_{l-1}} \to \mathbb{R}, 
X_l \mapsto F_l(X_l) := F(X_1,...,X_k), 
\end{equation}
then
\begin{equation}
\mbox{D}_{X_l}F(X_1,...,X_k) = \mbox{D}F_l(X_l)
\end{equation}
and
\begin{equation}
\nabla_{X_l}F(X_1,...,X_k) = \nabla F_l(X_l).
\end{equation}

\newpage
\section{Essential properties of Kronecker and Hadamard matrix products}\label{sec:appendix_2}
\setcounter{equation}{0}
\noindent For two matrices $A \in \mathbb{R}^{m \times n}$ and $B \in \mathbb{R}^{p \times q}$, the Kronecker matrix product is defined as
\begin{equation}
A \otimes B =
\begin{pmatrix}
a_{ij} \cdot B
\end{pmatrix}_{1 \le i \le m, 1 \le j \le n} =
\begin{pmatrix}
a_{11}B &	\cdots 	& a_{1n}B 		\\
\vdots 	&	\ddots 	& \vdots		\\
a_{m1}B	&	\cdots 	& a_{mn}B		
\end{pmatrix}
\in \mathbb{R}^{mp \times nq}.
\end{equation}
The Kronecker matrix product has the following properties.
\begin{itemize}[leftmargin = .7cm]
\item[(1)] For $A \in \mathbb{R}^{m \times n}$ and $B \in \mathbb{R}^{p \times q}$,
\begin{equation}\label{eq:kron_transposition}		
(A \otimes B)^T  = A^T \otimes B^T.
\end{equation}
\item[(2)] For $A \in \mathbb{R}^{m \times n}$, $B \in \mathbb{R}^{p \times q}$, $C \in \mathbb{R}^{n\times k}$, and $D \in \mathbb{R}^{q\times r}$,
\begin{equation}\label{eq:kron_mixed}		
(A \otimes B) \cdot (C \otimes D) = (AC) \otimes (BD).
\end{equation}
\item[(3)] For $A \in \mathbb{R}^{m \times n}$, $B \in \mathbb{R}^{p \times q}$, and $X \in \mathbb{R}^{n \times p}$,
\begin{equation}\label{eq:kron_vectorization}		
\vex(AXB) = (B^T \otimes A)\vex(X).
\end{equation}
\end{itemize}
For two matrices $A,B \in \mathbb{R}^{m \times n}$, the Hadamard matrix product is defined as
\begin{equation}
A \circ B =
\begin{pmatrix}
a_{ij} \cdot b_{ij}
\end{pmatrix}_{1 \le i \le m, 1 \le j \le n} =
\begin{pmatrix}
a_{11}b_{11} 	&	\cdots 		& a_{1n}b_{1n}	\\
\vdots 			&	\ddots 		& \vdots		\\
a_{m1}b_{m1}	&	\cdots 		& a_{mn}b_{mn}		
\end{pmatrix}
\in \mathbb{R}^{m \times n}.
\end{equation}
For proofs of these properties, see Sections 4.2 and 4.3 of \citet{horn_topics_1994}. It is readily verified that if $v,w \in \mathbb{R}^n$ and $\mbox{diag}(v) \in \mathbb{R}^{n \times n}$ denotes the diagonal matrix comprising the components of $v$ along its main diagonal, then
\begin{equation}\label{eq:hadamard_diagonal_matrix}
\mbox{diag}(v)w = v \circ w = w \circ v.
\end{equation} 

\newpage
\section{Jacobian matrices of essential neural network functions}\label{sec:appendix_3}
\setcounter{equation}{0}
\subsection*{Weight matrix-variate potential functions}
For $a \in \mathbb{R}^n$, let
\begin{equation}
\Phi_a : \mathbb{R}^{m \times (n+1)} \to \mathbb{R}^m,
W \mapsto \Phi_a(W) 
:= W
\cdot
\begin{pmatrix}
a \\ 1
\end{pmatrix} 
\end{equation}
denote a weight matrix-variate potential function (cf. eq. \eqref{eq:Phi_a}). Then the component functions of 
\begin{equation}
\vex(\Phi_a(W)) = \Phi_a(W)
\end{equation}
are given by
\begin{equation}
f_i : \mathbb{R}^{m \times (n+1)} \to \mathbb{R},
W \mapsto f_i(W) := \sum_{j = 1}^n w_{ij}a_j + w_{i,n+1}
\mbox{ for } i = 1,...,m.
\end{equation} 
The partial derivatives of the component functions $f_i, i = 1,...,m$ with respect to the elements $w_{kl}, k = 1,...,m, l = 1,...,n+1$ of $W$ are given by
\begin{equation}
\frac{\partial}{\partial w_{kl}}f_i(a) = 
\frac{\partial}{\partial w_{kl}} \sum_{j = 1}^n w_{ij}a_j + w_{i,n+1} 
= \begin{cases}
a_j & \mbox{ if  } k = i 	\mbox{ and } l = j \\
0   & \mbox{ if  } k \neq i \mbox{ or } l \neq j
\end{cases}
\end{equation}
for $k = 1,...,m$ and  $l = 1,...,n$, as well as by
\begin{equation}
\frac{\partial}{\partial w_{kl}}f_i(a) = 
\frac{\partial}{\partial w_{kl}} \sum_{j = 1}^n w_{ij}a_j + w_{i,n+1} 
= \begin{cases}
1 	& \mbox{ if  } k = i 	\quad\quad\quad\quad\quad\quad \\
0   & \mbox{ if  } k \neq i 
\end{cases}
\end{equation}
for $k = 1,...,m$ and  $l = n+1$. With eq. \eqref{eq:jacobian_f}, the Jacobian matrix of $\Phi_a$ at $W$ thus evaluates to

\begin{small}
\begin{align*}
\renewcommand{\arraystretch}{1.7}
\setcounter{MaxMatrixCols}{20}
\setlength\arraycolsep{2pt}
\begin{split}
& \mbox{D}\Phi_a(W) 
\\
\\
& := \left(\frac{\partial}{\partial (\vex(W))}_j f_i(W) \right)_{1 \le i \le m, 1 \le j \le m(n+1)} 
\\
\\
& = 
\begin{tiny}
\begin{pmatrix}
  \frac{\partial}{\partial w_{11}} 	  f_1(W)
& \cdots
& \frac{\partial}{\partial w_{m1}} 	  f_1(W)
& \frac{\partial}{\partial w_{12}} 	  f_1(W)
& \cdots
& \frac{\partial}{\partial w_{m2}} 	  f_1(W)
&  
& \frac{\partial}{\partial w_{1n}}    f_1(W)
& \cdots
& \frac{\partial}{\partial w_{mn}}    f_1(W)
& \frac{\partial}{\partial w_{1,n+1}} f_1(W)
& \cdots
& \frac{\partial}{\partial w_{m,n+1}} f_1(W)
\\
  \frac{\partial}{\partial w_{11}} 	  f_2(W)
& \cdots
& \frac{\partial}{\partial w_{m1}} 	  f_2(W)
& \frac{\partial}{\partial w_{12}} 	  f_2(W)
& \cdots
& \frac{\partial}{\partial w_{m2}} 	  f_2(W)
&  
& \frac{\partial}{\partial w_{1n}}    f_2(W)
& \cdots
& \frac{\partial}{\partial w_{mn}}    f_2(W)
& \frac{\partial}{\partial w_{1,n+1}} f_2(W)
& \cdots
& \frac{\partial}{\partial w_{m,n+1}} f_2(W)
\\
  \vdots
& \ddots
& \vdots
& \vdots
& \ddots
& \vdots
& \cdots
& \vdots
& \ddots
& \vdots
& \vdots
& \ddots
& \vdots
\\
  \frac{\partial}{\partial w_{11}} 	  f_m(W)
& \cdots
& \frac{\partial}{\partial w_{m1}} 	  f_m(W)
& \frac{\partial}{\partial w_{12}} 	  f_m(W)
& \cdots
& \frac{\partial}{\partial w_{m2}} 	  f_m(W)
& 
& \frac{\partial}{\partial w_{1n}}    f_m(W)
&  \cdots
& \frac{\partial}{\partial w_{mn}}    f_m(W)
& \frac{\partial}{\partial w_{1,n+1}} f_m(W)
& \cdots
& \frac{\partial}{\partial w_{m,n+1}} f_m(W)
\end{pmatrix}
\end{tiny}
\\
\\
& = 
\begin{pmatrix}
  a_1
& \cdots
& 0
& a_2
& \cdots
& 0
&  
& a_n
& \cdots
& 0
& 1
& \cdots
& 0
\\
  \vdots
& \ddots
& \vdots
& \vdots
& \ddots
& \vdots
&  \cdots
& \vdots
& \ddots
& \vdots
& \vdots
& \ddots
& \vdots
\\
  0
& \cdots
& a_1
& 0
& \cdots
& a_2
& 
& 0
& \cdots
& a_n
& 0
& \cdots
& 1
\end{pmatrix}
\\
\\
& = 
\begin{pmatrix}
  a_1 I_m 
& a_2 I_m 
& \cdots 
& a_n I_m 
& 1 I_m
\end{pmatrix}
\\
\\
& = (a^T \,\, 1) \otimes I_m.
\end{split}
\end{align*}
\end{small}
Note that $\mbox{D}\Phi_a(W)$ is an $m \times m(n+1)$ dimensional matrix.

\subsection*{Potential functions}
For $W \in \mathbb{R}^{m \times (n+1)}$, let  
\begin{equation}
\Phi_W : \mathbb{R}^n \to \mathbb{R}^m,
a \mapsto \Phi_W(a) 
:= W
\cdot
\begin{pmatrix}
a \\ 1
\end{pmatrix} 
\end{equation}
denote a potential function (cf. eq. \eqref{eq:Phi_W}). The component functions of $\Phi_W$ are given by
\begin{equation}
f_i : \mathbb{R}^n \to \mathbb{R}, a \mapsto f_i(a) := \sum_{j=1}^n w_{ij} a_j + w_{j,n+1}
\mbox{ for } i = 1,...,m.
\end{equation}
The partial derivatives of the component functions $f_i, i = 1,...,m$ with respect to the elements $a_j, j = 1,...,n$ of $a$ are given by
\begin{equation}
\frac{\partial}{\partial a_j}f_i(a) = 
\frac{\partial}{\partial a_j}\left(\sum_{j=1}^n w_{ij} a_j + w_{j,n+1}\right) = 
w_{ij} 
\end{equation}
for $i = 1,...,m$ and $j = 1,...,n$. With eq. \eqref{eq:jacobian_f}, we thus have for the Jacobian matrix of $\Phi_W$ at $a$
\begin{align}
\begin{split}
\mbox{D}\Phi_W(a) 
:= \left(\frac{\partial}{\partial a_j}f_i(a) \right)_{1 \le i \le m, 1 \le j \le n}
=  \left(w_{ij} \right)_{1 \le i \le m, 1 \le j \le n}
:= W_{\bullet}, 
\end{split}
\end{align}
where $W_{\bullet} \in \mathbb{R}^{m \times n}$ is defined as the matrix resulting from removing the last column of the weight matrix $W \in \mathbb{R}^{m \times (n+1)}$.

\subsection*{Component-wise activation functions} 
For an activation function 
\begin{equation}
\sigma : \mathbb{R} \to \mathbb{R}, z_i \mapsto \sigma(z_i),
\end{equation}
let
\begin{equation}
\Sigma : \mathbb{R}^n \to \mathbb{R}^n, 
z \mapsto \Sigma(z) := (\sigma(z_1), ..., \sigma(z_n))^T
\end{equation}
denote a component-wise activation function (cf. eq. \eqref{eq:Sigma}). The component functions of $\Sigma$ are given by
\begin{equation}
f_i : \mathbb{R}^n \to \mathbb{R}, z \mapsto f_i(z) := \sigma(z_i)
\mbox{ for } i = 1,...,n.
\end{equation}
The partial derivatives of the component functions $f_i, i = 1,...,n$ with respect to the elements $z_j, j = 1,...,n$ of $z$ are given by
\begin{equation}
\frac{\partial}{\partial z_j}f_i(z)
= \frac{\partial}{\partial z_j}\sigma(z_i)
= 
\begin{cases}
\sigma'(z_i) & \mbox{ if } i = j	\\
0 			 & \mbox{ if } i \neq j
\end{cases}.
\end{equation}
With eq. \eqref{eq:jacobian_f}, we thus have for the Jacobian matrix of $\Sigma$ at $z$
\begin{align}
\begin{small}
\mbox{D}\Sigma(z) 
:= \left(\frac{\partial}{\partial z_j}f_i(z) \right)_{1 \le i \le n, 1 \le j \le n}
 = \left(
\begin{cases}
\sigma'(z_i) & \mbox{ if } i = j	\\
0 			 & \mbox{ if } i \neq j
\end{cases} 
\right)_{1 \le i \le n, 1 \le j \le n}
= \mbox{diag}(\sigma(z_1), ..., \sigma(z_n)). 
\end{small}
\end{align}
Note that $\mbox{D}\Sigma(z)$ is an $n\times n$-dimensional matrix.

\end{document}